\documentclass[english]{article}
\usepackage[latin9]{inputenc}
\usepackage{geometry}
\usepackage{array}
\usepackage{float}
\usepackage{amssymb}
\usepackage{graphicx}
\usepackage{epstopdf}

\makeatletter

\providecommand{\tabularnewline}{\\}
\newcommand{\lyxdot}{.}

\floatstyle{ruled}
\newfloat{algorithm}{tbp}{loa}
\providecommand{\algorithmname}{Algorithm}
\floatname{algorithm}{\protect\algorithmname}

\usepackage{times}
\newtheorem{proposition}{Proposition}
\usepackage{epstopdf}

\makeatother

\usepackage{babel}
\begin{document}

\title{Tree-based iterated local search for Markov random fields with applications in image analysis}

\author{Truyen Tran, Dinh Phung, Svetha Venkatesh \\
Center for Pattern Recognition and Data Analytics, Deakin University \\  	
Waurn Ponds, VIC 3216, Australia.\\
\{truyen.tran,dinh.phung,svetha.venkatesh\}@deakin.edu.au}

\date{}

\maketitle

\begin{abstract}
The \emph{maximum a posteriori} (MAP) assignment for general structure
Markov random fields (MRFs) is computationally intractable. In this
paper, we exploit tree-based methods to efficiently address this problem.
Our novel method, named Tree-based Iterated Local Search (T-ILS) takes
advantage of the tractability of tree-structures embedded within MRFs
to derive strong local search in an ILS framework. The method efficiently
explores exponentially large neighborhood and does so with limited
memory without any requirement on the cost functions. We evaluate
the T-ILS in a simulation of Ising model and two real-world problems
in computer vision: stereo matching, image denoising. Experimental
results demonstrate that our methods are competitive against state-of-the-art
rivals with a significant computational gain.
\end{abstract}
\global\long\def\xb{\mathbf{x}}
\global\long\def\x{\mathbf{x}}

\global\long\def\yb{\mathbf{y}}
\global\long\def\y{\mathbf{y}}

\global\long\def\vertices{\mathcal{V}}

\global\long\def\edges{\mathcal{E}}

\global\long\def\neighbours{\mathcal{N}}

\section{Introduction}

Markov random fields (MRFs) \cite{Besag-74,Geman-Geman84,Lauritzen96}
are popular probabilistic graphical representation for encoding the
object's state space using sites and the correlation between local
objects using edges. For example, a grid is a powerful image representation
for pixels, where each site represents a pixel state (e.g., intensity
or label) and an edge represents local relations between neighboring
pixels (e.g., smoothness). In other words, we can encode prior knowledge
about the nature of interaction between objects in a MRF. Once the
model has been built, we can perform inference in a principled manner.

One of the most important MRF inference problems in many applications
is finding \emph{maximum a posteriori} (MAP) assignment. The goal
is to search for the the most probable state configuration of all
objects, or equivalently, the lowest energy of the model. The problem
is known to be NP-complete \cite{Boykov-et-al01}. Given a MRF of
$N$ objects each of which has $S$ states, a brute-force search must
explore $S^{N}$ state configurations. In image denoising, for example,
$N$ is the number of pixels, which can range from $10^{4}-10^{7}$,
and $S$ is the number of pixel intensity levels, which are in the
order of $2^{8}-2^{32}$. This calls for heuristic methods which can
find reasonable solutions in practical time.

There have been a great number of attempts to solve this problem.
An early approach was based on simulated annealing (SA), where the
convergence is also guaranteed for log-time cooling schedule \cite{Geman-Geman84}.
The main drawback of this approach is its low speed -- it takes long
time to achieve reasonably good solutions. Another early attempt involves
local greedy search, and the iterated conditional mode (ICM) \cite{Besag-86}
is probably the most well-known method. In probabilistic terms, it
iteratively seeks for the mode of the local distribution of states
of an object conditioned on the states of nearby objects. Translated
in combinatorial terms, it greedily searches for the local valley
- here the neighborhood size is limited to $S$ - the number of possible
states per object. Not surprisingly, this method is prone to getting
stuck at poor local minima.

A more successful approach is belief-propagation (BP) \cite{Pearl88},
which exploits the \emph{problem structure} better than the ICM. The
main idea is that we maintain a set of messages sending simultaneously
along all edges of the MRF network. A message carries the information
of the state of the sites it originates from, and as a result, any
update of a particular site is informed by messages from all nearby
sites. However, this method can only be guaranteed to work for a limited
class of network structures -- when the network reduces to a tree,
where there are no loops in the network. Another drawback is that
the memory requirement for BP is high - this is generally linear in
the number of edges in the network. More recently, efficient algorithms
with theoretical guarantees have been introduced based on the theory
of graph cuts \cite{Boykov-et-al01}. This class of algorithms, while
being useful in certain computer vision problems, has limitation in
the range of problems it can solve -- the energy formulation must
admits a certain \emph{metric} form \cite{Boykov-et-al01,szeliski2007comparative}.
In effect, these algorithms are not applicable to problems where energy
functions are estimated from data and their functional forms are not
known in priori.

Given this ground, it is desirable to have an approximate algorithm
which is fast enough (better than ICM - for example), consumes little
memory and does not have any specific requirements of network structures
or energy functionals. A meta-heuristic solution has already been
proposed in the combinatorial optimization literature - the Iterated
Local Search (ILS) (e.g. see \cite{Lourenco-et-al03}). Typically,
ILS encourages jumping between local minima, which can be found by
local search methods such as ICM. This algorithm, however, does not
exploit any model-specific information, and thus could be inefficient
for the MAP assignment in MRFs. To this end, we propose a novel algorithm
called \emph{Tree-based Iterated Local Search} (T-ILS), which combines
the strength of BP and ILS. This algorithm exploits the fact that
BP works efficiently on tree structures, in which the time complexity
is linear in the number of edges in the tree, and thus BP is an effective
candidate for locating good local minima. The main difference from
standard tree-based BP is that our trees are \emph{conditional} trees
which are built upon fixing states of neighbor leaves. When combined
with ILS, we have a heuristic algorithm that is less likely to get
stuck in poor local minima, and has better chance to reach high quality
solution, or even global minima.

We evaluate our tree-based inference methods on three benchmark problems:
Ising model, stereo correspondence and image denoising. We empirically
demonstrate that T-ILS attains good performance, while requiring less
training time and memory than the loopy BP, which is one the state-of-the-arts
for these problems. 

To summarize, our main contributions are the proposal and evaluation
of fast and lightweight tree-based inference methods in MRFs. Our
choice of trees on $N=W\times H$ images requires only $\mathcal{O}(2D)$
memory where $D=\max\{W,H\}$ and two passes over all sites in the
MRFs per iteration. These are much more economical than $\mathcal{O}(4WH)$
memory and many passes needed by the traditional loopy BP.

This paper is organized as follows. Section~\ref{sec:Problem-Setting}
describes MRFs, the MAP assignment problem, and belief propagation
for trees. In Section~\ref{sec:Iterated-Strong-Local}, the concept
of conditional trees is defined, followed by two algorithms: the strong
local search T-ICM and the global search T-ILS. Section~\ref{sec:exp}
provides empirical support for the performance of the T-ICM and T-ILS.
Finally, Section~\ref{sec:conclusion} concludes the paper.

\section{Related work \label{sec:Related-Work}}

The MAP assignment for Markov random fields (MRFs) as a combinatorial
search problem has attracted a great amount of research in the past
several decades, especially in the area of computer vision \cite{li1995markov}
and probabilistic artificial intelligence \cite{Pearl88}. The problem
is known to be NP-hard \cite{shimony1994finding}. For example, in
labeling of an image of size $W\times H$, the problem space is $S^{WH}$
large, where $S$ is the number of possible labels per pixel. 

Techniques for solving the MAP assignment can be broadly classified
into stochastic and deterministic classes. In early days, the first
stochastic algorithms were based on simulated annealing (SA) \cite{Kirkpatrick-et-al83}.
The first application of SA to Markov random fields (MRFs) with provable
convergence was perhaps the work of \cite{Geman-Geman84}. The main
drawback of this method is slow convergence toward good solutions
\cite{szeliski2007comparative}. Nature-inspired algorithms were also
popular, especially the family of genetic algorithms \cite{brown2002markov,kim1998mrf,kim2009markov,maulik2009medical,tseng1999genetic}.
Some attempts using ant colony optimization and tabu-search have also
been made \cite{ouadfel2003mrf,yousefi2012brain}.

Deterministic algorithms started in parallel with iterated conditional
model (ICM) \cite{Besag-86}. It is a simple greedy search strategy
that updates one label at a time. Thus it is very slow and sensitive
to bad initialization. A more successful approach is based on Pearl's
loopy belief propagation (BP) \cite{Pearl88}. Due to its nature of
using local information (called ``messages'') to update ``belief''
about the optimal solution, loopy BP is also called a \emph{message
passing} algorithm. Although loopy BP is neither guaranteed to converge
at all nor to reach global optima, empirical evidences so far have
indicated that it is very competitive against state-of-the-arts in
a variety of image analysis problems \cite{Felzenszwalb-Huttenlocher-IJCV06,szeliski2007comparative}.
Research on improving loopy BP is currently a very active topic in
a range of disciplines, from artificial intelligence, statistical
physics, computer visions to social network analysis \cite{duchi2007uco,Felzenszwalb-Huttenlocher-IJCV06,hazan2010norm,jojic2010accelerated,Kolmogorov-PAMI06,kumar2012message,meltzer2009convergent,zheng2012map,Wainwright-et-al-TR03}.
The most recent development centers around convex analysis \cite{johnson2007lrm,kumar2009analysis,ravikumar2006qpr,Wainwright-et-al05TIT,werner2007linear}.
In particular, the MAP is converted into linear programming with relaxed
constraints from which a mixture of convex optimization and message
passing can be used. Loopy BP also plays role in improving evolutionary
algorithms under probabilistic graphical model representation \cite{larranaga2012review}.

Another powerful class of algorithms in computer vision is graph cuts
\cite{Boykov-et-al01,szeliski2007comparative}. They are, nevertheless,
designed with specific cost functions in mind (i.e. \emph{metric}
and \emph{semi-metric}) \cite{kolmogorov2004energy}, and therefore
inapplicable for generic cost functions such as those resulting from
learning. Again, research in graph cuts is an active area in computer
vision \cite{bhusnurmath2008graph,boykov2004experimental,kohli2010dynamic,kolmogorov2007minimizing,kumar2009map,lempitsky2007logcut}.
Interestingly, it has been recently proved that graph cuts are in
fact loop BP \cite{tarlow2011graph}. 

It is fair to say that the deterministic approach has become dominant
due to their performance and theoretical guarantee for certain classes
of problems \cite{kappes2014comparative}. However, the problem is
still unsolved for general settings, thus motivates this paper. Our
approach has both the deterministic and heuristic nature. It relies
on the concept of \emph{strong local search} using the deterministic
method of BP, which is efficient in trees. The local search is strong
because it covers a significant number of sites, rather than just
one, which is often found in other local search methods such as ICM
\cite{Besag-86}. The neighborhood size in our method is very large
\cite{ahuja2002survey}. For typical image labeling problems, the
size is $S^{0.5WH}$ for an image of height $H$ and width $W$ and
label size of $S$. Standard local search like ICM only explores the
neighborhood of size $S$ at a time. Once a strong local minimum is
found, a stochastic procedure based on iterated local search \cite{Lourenco-et-al03}
is applied to escape from it and explores a better local minimum. 

The idea of exploiting the trees in MRF in image analysis is not entirely
new. In early days, a spanning tree is used to approximate the entire
MRF \cite{chow1968adp,Wu-Doerschuk-PAMI95}. This method is efficient
but the approximation quality may hurt because the number of edges
in tree is far less than that in the original MRF. Another way is
to built a hierarchical MRF with multiple resolutions \cite{Willsky-IEEE02},
but this is less applicable for flat image labeling problems. Our
method differs from these efforts in that we use trees embedded in
the original graph rather than building an approximate tree. Second,
our trees are conditional -- trees are defined on the values of its
leaves. Third, trees are selected as we go during the search process.

More recently, trees are used in variants of loop BP to specify the
orders which messages are scheduled \cite{Wainwright-et-al05TIT,sontag2009tree}.
Our method can also viewed along this line by differs in the way trees
are built and messages are updated. In particular, our trees are conditional
on neighbor labeling, which is equivalent to collapsing an associated
message to a single value.

Iterated Local Search (ILS), also known as basin hopping \cite{wales1997global},
has been used in related applications such as image registration \cite{cordon2006image},
and structure learning in probabilistic graphical models \cite{biba2008discriminative}.
The success of ILS depends critically on the local search and the
perturbation (basin hopping) strategy \cite{lourencco2010iterated}.
In \cite{wales1997global}, for example, a powerful local search based
on conjugate gradients is essential to reach global solution for the
Lennard-Jones clusters problem. Our work builds strong local search
using tree-based BP on the discrete spaces rather than continuous
ones.

\section{Markov random fields for image labeling \label{sec:Problem-Setting}}

In this section we introduce Markov random fields (MRFs) and the MAP
assignment problem with application to image labeling. A MRF is a
probabilistic way to express the uncertainty in the discrete system
of many interacting variables each of which is characterized by a
set of possible states \cite{Pearl88}. In what follows, we briefly
describe the MRF and its MAP assignment problem and focus on the minimization
of model energy.

Formally, a MRF specifies a random field $\xb=\{x_{i}\}_{i=1}^{N}$
over a graph $\mathcal{G}=(\mathcal{V},\mathcal{E})$, where $\mathcal{V}$
is the collection of $N$ sites $\{i\}$, $\mathcal{E}$ is the collection
of edges $\{(i,j)\}$ between sites, and $x_{i}\in L$ represents
states at site $i$. The entire system is random because the uncertainty
in specifying the exact state of each site. One of the main objectives
is to compute the most probable specification, also known as MAP assignment,
which is the main focus of the paper.

\subsection{Image labeling as energy minimization and MAP assignment}

In image labeling problem, an image $\yb$ is a collection of pixels
arranged in a particular geometrical way, as defined by the graph
$\mathcal{G}$. Typically, we assume a grid structure over pixels,
where every inner pixel has exactly four nearby pixels. A labeling
$\xb$ is the assignment of each pixel $y_{i}$ a corresponding label
$x_{i}$ for all $i=1,2,...,N$. 

A full specification of an MRF over the labeling $\xb$ given the
image $\yb$ can be characterized by its energy. Assuming pairwise
interaction between connected sites, the energy is the sum of local
energies as follows:

\begin{eqnarray}
E(\xb,\yb) & = & \sum_{i\in\mathcal{V}}E_{i}(x_{i},\yb)+\sum_{(i,j)\in\mathcal{E}}E_{ij}(x_{i},x_{j})\label{eq:energy-decompose}
\end{eqnarray}
The singleton energy $E_{i}(x_{i},\yb)$ encodes the disassociation
between the label $x_{i}$ and the features of image $\yb$ at site
$i$. In image denoising, for example, where $y_{i}$ is the corrupted
pixel and $x_{i}$ is the true pixel we may use $E_{i}(x_{i},\yb)=\left|x_{i}-y_{i}\right|$
as the cost due to corruption. The pairwise energy $E_{ij}(x_{i},x_{j})$
often captures the smoothness nature of image, that is, two nearby
pixels tend to be similar. For example, $E_{ij}(x_{i},x_{j})=\lambda\left|x_{i}-x_{j}\right|$
is a cost of difference between two labels, where $\lambda>0$ is
a problem-specific parameter. 

In image labeling, the task is to find the optimal $\xb^{map}$ that
minimizes the energy $E(\xb,\yb)$, which now plays the role of the
\emph{cost function:}
\begin{eqnarray}
\xb^{map} & = & \arg\min_{\xb}E(\xb,\yb)\label{eq:MAP}
\end{eqnarray}
For example, in image denoising, this translates to finding a map
of intensity that admits both the low cost of corruption and high
smoothness.

The formal justification for the energy minimization in MRF can be
found through the probability of the labeling defined as

\begin{equation}
P(\xb\mid\yb)=\frac{1}{Z(\yb)}e^{-E(\xb,\yb)}\label{eq:Boltzmann-distribution}
\end{equation}
where $Z(\yb)$ is the normalization term. Thus minimizing the energy
is equivalent to finding the most probable labeling $\xb^{map}$.
As $P(\xb\mid\yb)$ is often called the \emph{posterior} distribution
in computer vision%
\footnote{The term \emph{posterior} comes from the early practice in computer
vision in which $P(\yb\mid\xb)$ is first defined then linked to $P(\xb\mid\yb)$
through the Bayes rule:

\[
P(\xb\mid\yb)=\frac{P(\xb)P(\yb\mid\xb)}{P(\yb)}
\]
where $P(\xb)$ is called the \emph{priori}. However in this paper
we will work directly with $P(\xb\mid\yb)$ for simplicity. The posterior
is recently called conditional random fields in machine learning \cite{lafferty01conditional,truyen-thesis08}%
}, the energy minimization problem is also referred to as \emph{maximum
a posteriori} (MAP) assignment.

\subsection{Local search: Iterated conditional mode \label{sub:Iterated-Conditional-Mode}}

Iterated conditional mode (ICM) \cite{Besag-86} is a simple local
search algorithm. It iteratively finds the local optimal labeling
for each site $i$ as follows
\begin{eqnarray}
x_{i}^{*} & =\arg\min_{x_{i}} & \left\{ E_{i}(x_{i},\yb)+\sum_{j\in\neighbours(i)}E_{ij}(x_{i},x_{j})\right\} \label{eq:ICM-energy}
\end{eqnarray}
where $\neighbours(i)$ is the set of sites connected to the site
$i$, often referred to as \emph{Markov blanket}. The Markov blanket
shields a site from the long-range interactions with remote sites,
due to a special property of MRF known as \emph{Markov property,}
which states that probability of a label assignment at site $i$ given
all other assignments depends only on the nearby assignments \cite{Hammersley-Clifford71,Lauritzen96}.
The probabilistic interpretation of Eq.~(\ref{eq:ICM-energy}) is

\[
x_{i}^{*}=\arg\max_{x_{i}}P\left(x_{i}\mid\{x_{j}\}_{j\in\neighbours(i)},\y\right)
\]
The local update in Eq.~(\ref{eq:ICM-energy}) is repeated for all
sites until no more improvement can be made. This procedure is guaranteed
to find a local minimum energy in a finite number of steps. However,
the solutions found by the ICM are sensitive to initialization and
often unsatisfactory for image labeling \cite{szeliski2007comparative}.

\subsection{Exact global search on trees: Belief-propagation \label{sec:Belief-Propagation}}

\begin{figure}[htb]
\begin{centering}
\begin{tabular}{c}
\includegraphics[width=0.7\textwidth]{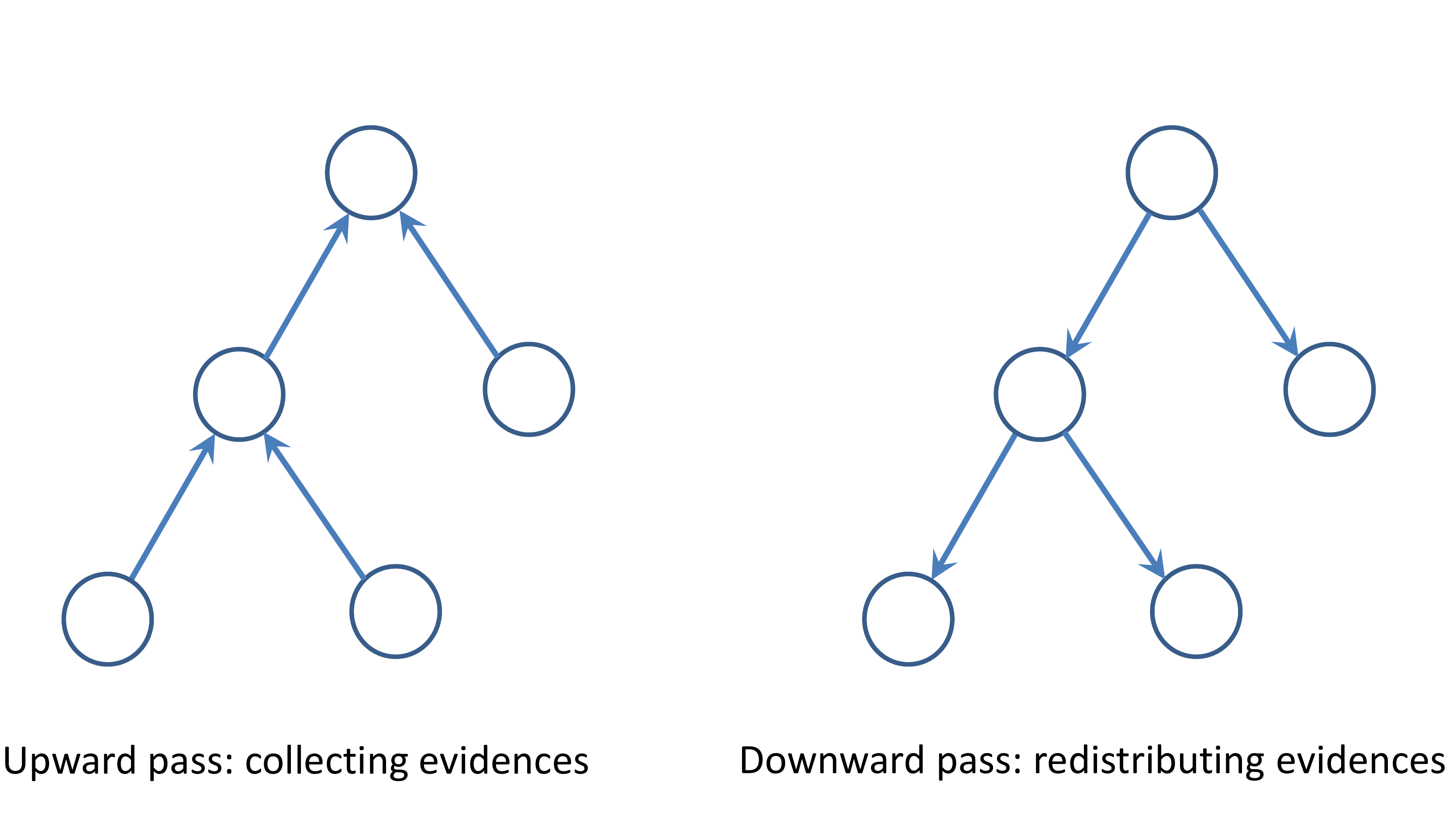}\tabularnewline
\end{tabular}
\par\end{centering}

\caption{Belief propagation on trees: the two-pass procedure.\label{fig:The-two-pass-procedure:}}

\label{fig:bgr-mrf-tree-2pass} 
\end{figure}

Belief-propagation (BP) was first proposed as an inference method
on MRF with tree-like structures in the field of artificial intelligence
\cite{Pearl88}. BP operates by sending \emph{messages} between connecting
sites. For this reason, it is also called \emph{message passing} algorithm.
It is efficient because instead of dealing with all the sites simultaneously,
we need to compute messages passing between two local sites. At each
site, local information modifies the incoming messages before sending
out to neighbor sites.

\paragraph{General BP.}

The general rule is that the message sent from site $j$ to site $i$
in the tree is computed as follows 
\begin{eqnarray}
\mu_{j\rightarrow i}(x_{i}) & = & \min_{x_{j}}\left(E_{j}(x_{j},\yb)+E_{ij}(x_{i},x_{j})+\sum_{k\in\neighbours(j),k\ne i}\mu_{k\rightarrow j}(x_{j})\right)\label{bgr-BP-update}
\end{eqnarray}
That is, the outgoing message is aggregated from all incoming messages,
except for the one in the opposite direction. Messages can be initialized
arbitrarily, and the procedure is guaranteed to stop after finite
steps on trees \cite{Pearl88}. The optimal labeling could be estimated
as follows
\begin{equation}
x_{i}^{map}=\arg\min_{x_{i}}\left(E_{i}(x_{i},\yb)+\sum_{k\in\neighbours(i)}\mu_{k\rightarrow i}(x_{i})\right)\label{eq:BP-max-marginal}
\end{equation}
It has been proved that the labeling obtained in this way is indeed
globally optimal \cite{Pearl88}, i.e.,

\[
\xb^{map}=\arg\min_{\xb}E(\xb,\yb)
\]

\paragraph{2-pass BP.}

A more efficient variant of BP is the 2-pass procedure, as summarized
in Fig.~\ref{fig:The-two-pass-procedure:}. First we pick one particular
site as the root. Since the graph has no loops, there is a single
path from a site to any other sites in the graph, and each site, except
for the root, has exactly one parent. The procedure consists of two
passes: 
\begin{itemize}
\item \emph{Upward pass}: Messages are first initiated at the leaves, and
are set to $0$. Then all messages are sent upward and updated as
messages converging at common parents along the paths from leaves
to the root. The pass stops when all the messages reach the root. 
\item \emph{Downward pass}: Messages are combined and re-distributed downward
from the root back to the leaves. The messages are then terminated
at the leaves. 
\end{itemize}
The 2-pass BP procedure is a remarkable algorithm: it searches through
the combinatorial space of $S^{N}$ using only $\mathcal{O}\left(2NS^{2}\right)$
operations and $\mathcal{O}\left(2NS\right)$ memory to store all
the messages.

\paragraph{Remark.}

We note in passing that this procedure may be known as \emph{min-sum},
\emph{max-product}, or simply BP. The term max-product comes from
the fact that we can operate in the potential domain by taking the
exponential of negative energies in Eqs.~(\ref{bgr-BP-update},\ref{eq:BP-max-marginal}),
and turn mins into maxes and sums into products.

\subsection{Approximate global search on general graphs: Loopy belief-propagation
\label{sec:Loopy-BP}}

Standard MRFs in image analysis are usually not tree-structured. A
common topology is a grid in which each site represents a pixel and
has four neighbors. Thus the resulting graph has many cycles, rendering
the standard 2-pass BP algorithm useless. 

However, an approximation to exact BP has been suggested. Using the
general BP described above, messages are sent across all edges without
worrying about the order \cite{Pearl88}. At each step, the messages
are updated using Eq.~(\ref{bgr-BP-update}). After some stopping
criteria are met, we still use Eq.~(\ref{eq:BP-max-marginal}) to
find the best labeling. This procedure is often called \emph{loopy
BP} due to the presence of loops in the graph. The heuristic has been
shown to be useful in several applications \cite{Murphy-et-al-UAI99}
and this has triggered much research on improving it \cite{duchi2007uco,Felzenszwalb-Huttenlocher-IJCV06,kolmogorov2005otr,Sun-et-alPAMI03,sanghavi2007elr,Wainwright-et-al05TIT,Wainwright-et-al04,weiss2001osm,yanover2003fmm,yanover2006lpr}.

The main drawback of loopy BP is lack of convergence guarantee for
general problems. In our simulation of Ising models (Sec.~\ref{sub:Simulated-Ising-model}),
loop BP clearly fails in the cases where interaction energies dominate
singleton energies, that is $\left|E_{ij}(x_{i},x_{j})\right|\gg\left|E_{i}(x_{i},\yb)\right|$
for all $i,j$ (see Fig.~\ref{fig:Ising}). Another drawback is that
the memory will be very demanding for large images. For grid-image,
the memory needed is $\mathcal{O}\left(4HWS\right)$ which in the
order of gigabytes, and thus may not be suitable for devices with
small footprint.

\section{Iterated strong local search \label{sec:Iterated-Strong-Local}}

In this section we present a method to exploit the efficiency of the
BP on trees to build strong local search for the MAP assignment problem
in Markov random fields. By `strong', we mean the quality of the local
solution found by the procedure is often much better than standard
greedy local search. Although a typical MRF in computer vision is
not a tree, we observe that it can be thought as a super-imposition
of trees. Second, due to the Markov property of MRFs, described in
Sec.~\ref{sub:Iterated-Conditional-Mode}, variables in a tree can
be shielded from other variables through the Markov blanket of the
tree. This gives rise to the concept of \emph{conditional trees},
which we present subsequently.

\subsection{Conditional trees \label{sec:Conditional-trees} }

\begin{figure}[htb]

\begin{centering}
\begin{tabular}{cc}
\includegraphics[width=0.45\textwidth]{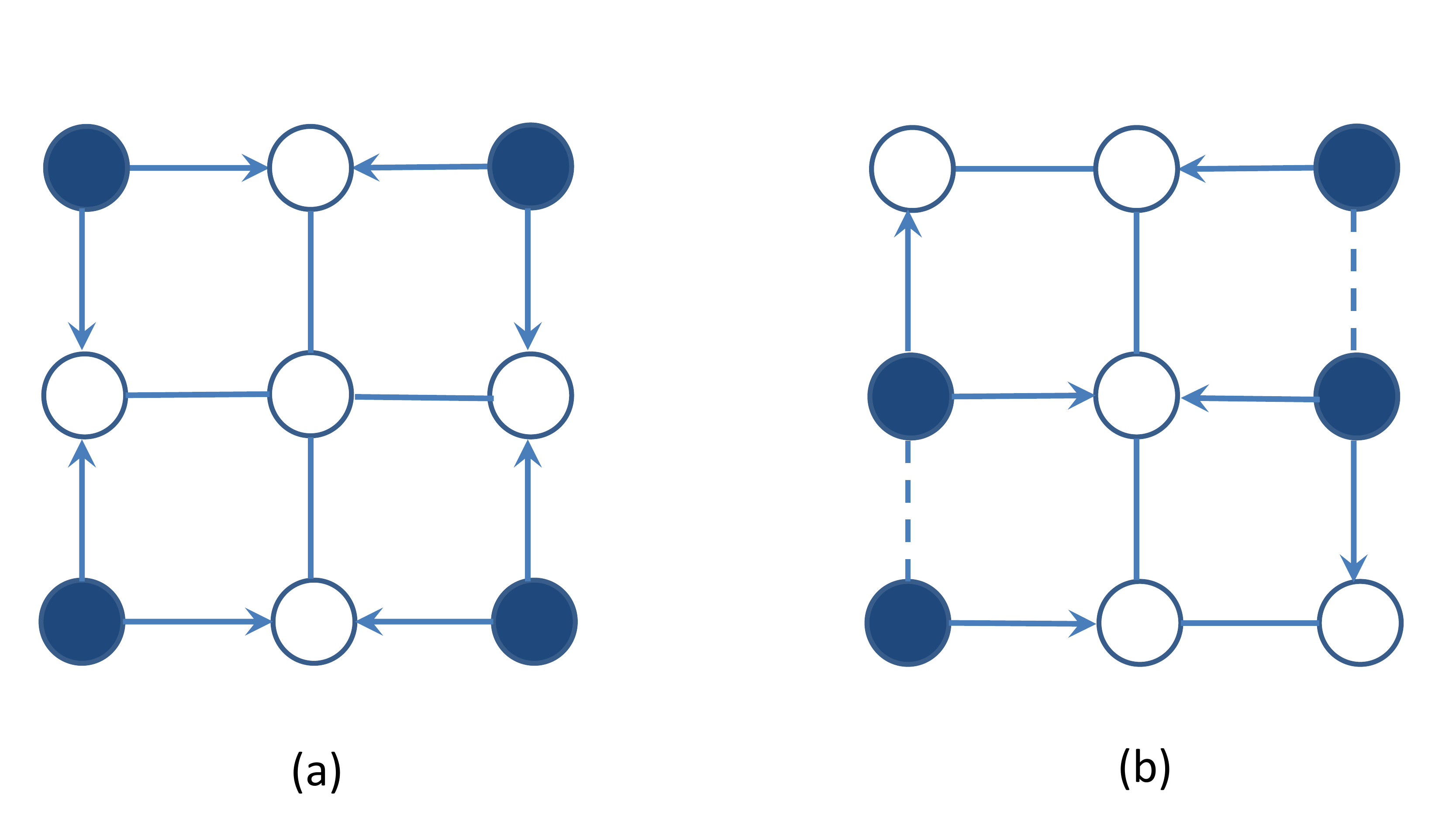}  & \includegraphics[width=0.45\textwidth]{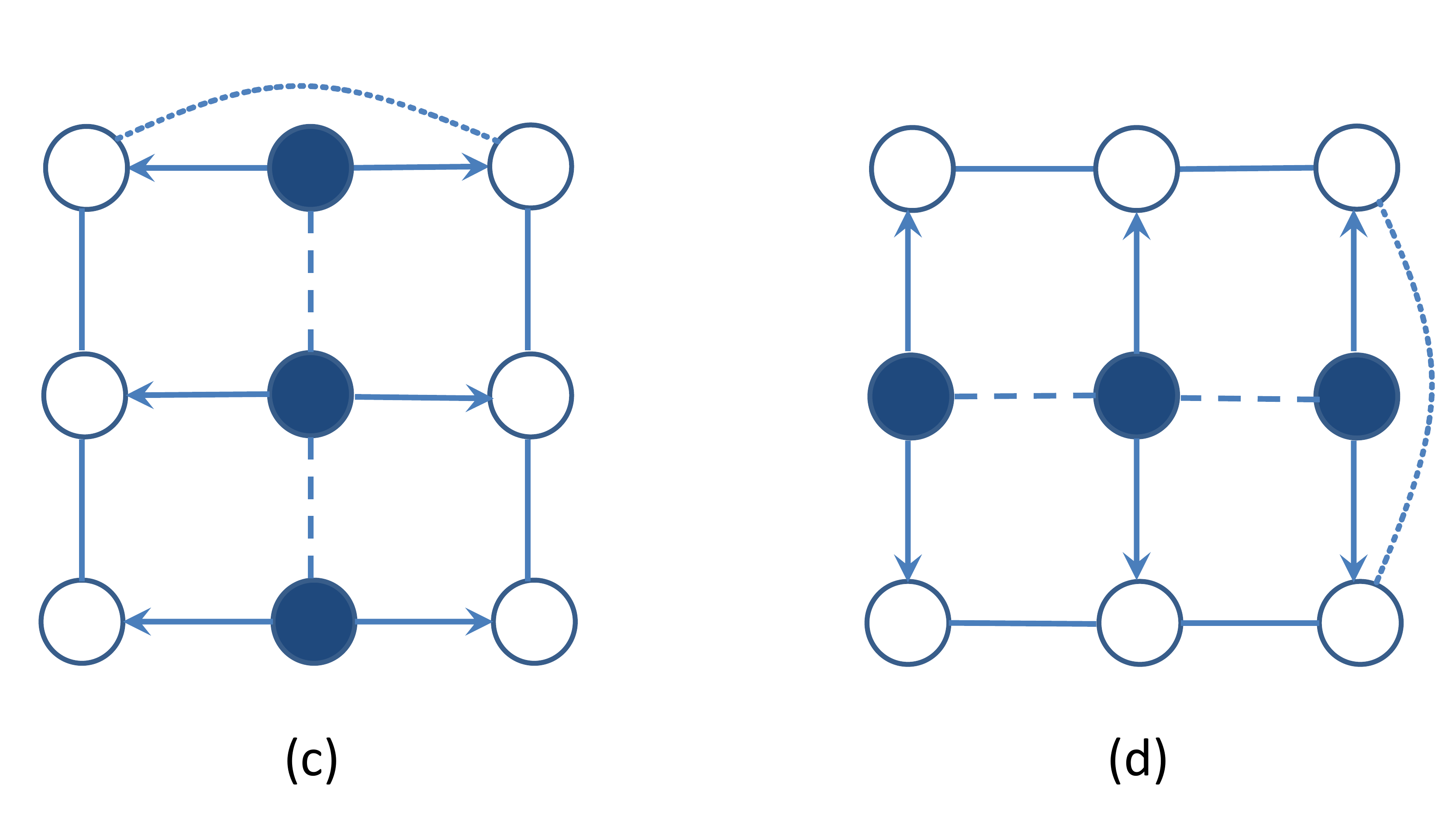}\tabularnewline
\end{tabular}
\par\end{centering}

\caption{Examples of conditional trees on grid (connecting empty sites). Filled
nodes are labeled sites. Arrows indicate absorbing direction, dashes
represent unused interactions. Dotted lines in (c,d) are dummy edge
(with zeros interacting energy) that connects separate sub-trees together
to form a full tree. \label{fig:conditional}}
\end{figure}

For concreteness, let us consider grid-structured MRFs. There are
more than one way to extract a tree out of the grid, as shown in Fig~\ref{fig:conditional}.
In particular, we fix some labeling to some sites, leaving the rest
forming a tree. Consider a tree $\tau$ and let $\xb_{\tau}=\{x_{i}\mid i\in\tau\}$,
and $\xb_{\neg\tau}=\{x_{i}\mid i\notin\tau\}$. Denote by $\neighbours(\tau)$
the set of sites connecting to $\tau$ but do not belong to $\tau$.
This is essentially the Markov blanket of the tree $\tau$. Thus the
collection of sites $\left(\tau,\neighbours(\tau)\right)$ and the
partial labeling of the neighbor sites $\xb_{\neighbours(\tau),}$
forms a conditional tree. The energy of the conditional tree can be
written as: 

\begin{eqnarray}
E_{\tau}\left(\x_{\tau},\xb_{\neighbours(\tau)},\y\right) & = & \sum_{i\in\tau}E_{i}^{*}(x_{i},\yb)+\sum_{i,j\in\tau}E_{ij}(x_{i},x_{j})\label{eq:cond-tree-energy}
\end{eqnarray}
where
\begin{equation}
E_{i}^{*}(x_{i},\yb)=E_{i}(x_{i},\yb)+\sum_{(i,j)\in\mathcal{E},j\in\neighbours(\tau)}E_{ij}(x_{i},x_{j})\label{eq:absorbing}
\end{equation}
In other words, the interacting energies at the tree border are \emph{absorbed}
into the singleton energy of the ordering sites. In Fig.~\ref{fig:conditional}
the absorbing direction is represented by an arrow.

The minimizer of this conditional tree energy can be found efficiently
using the BP:
\begin{equation}
\hat{\x}_{\tau}=\arg\min_{\x_{\tau}}E_{\tau}\left(\x_{\tau},\xb_{\neighbours(\tau)},\y\right)\label{eq:MAP-cond-tree-energy}
\end{equation}
This is due to the fact that Eq.~(\ref{eq:cond-tree-energy}) now
has the form of Eq.~(\ref{eq:energy-decompose}). 

One may wonder how the minima of the energy of conditional trees $E_{\tau}\left(\x_{\tau},\xb_{\neighbours(\tau)},\y\right)$
relate to the minima of the entire systems $E\left(\x,\y\right)$.
We present here two theoretical results of this connection. First,
the local minimum found by Eq.~(\ref{eq:MAP-cond-tree-energy}) is
also a local minimum of $E\left(\x,\y\right)$:

\begin{proposition} \label{prop:T-ICM-local-min}

Finding the mode $\hat{\xb}_{\tau}$ as in Eq.~(\ref{eq:MAP-cond-tree-energy})
guarantees a local minimization of model energy over all possible
tree labelings. That is

\[
E(\hat{\xb}_{\tau},\x_{\neg\tau},\y)\le E(\xb_{\tau},\x_{\neg\tau},\y)
\]
for all $\x_{\tau}\ne\hat{\x}_{\tau}$.

\end{proposition}

\paragraph{Proof:}

The proof is presented in Appendix~\ref{sub:Proof-of-T-ICM-local-opt}.

The second theoretical result is that the local minimum found by Eq.~(\ref{eq:MAP-cond-tree-energy})
is indeed the global minimum of the entire system if all other labels
outside the tree happen to be part of the optimal labeling:

\begin{proposition} \label{prop:MAP-property}

If $\x_{\neg\tau}\in\x^{map}$ then $\hat{\x}{}_{\tau}\in\x^{map}$. 

\end{proposition}

\paragraph{Proof:}

We first observe that since $E(\x^{map},\y)$ is the lowest energy
then 
\begin{eqnarray}
E(\x^{map},\y) & \le & E(\hat{\x}_{\tau},\x_{\neg\tau},\y)\label{E-upperboud}
\end{eqnarray}
Now assume that $\hat{\x}_{\tau}\notin\x^{map}$, so there must exist
$\x'_{\tau}\in\x^{map}$ that $\hat{\x}_{\tau}\ne\x'_{\tau}$ and
$E(\x'_{\tau},\x_{\neg\tau},\y)>E(\hat{\x}_{\tau},\x_{\neg\tau},\y)$,
or equivalently $E(\x^{map},\y)>E(\hat{\x}_{\tau},\x_{\neg\tau},\y)$,
which contradicts with Eq.~(\ref{E-upperboud}) $\blacksquare$

The derivation in Eq.~(\ref{eq:cond-tree-energy}) from the probabilistic
formulation is presented in Appendix.~\ref{sub:Distribution-over-conditional}.

\subsection{Tree-based ICM (T-ICM): conditional trees for strong local search
\label{sub:Tree-based-ICM-(T-ICM)}}

As conditional trees can be efficient to estimate the optimal labeling,
we propose a method in the spirit of the simple local search ICM \cite{Besag-86}
(Section~\ref{sub:Iterated-Conditional-Mode}). First of all, a set
of conditional trees $\mathcal{T}$ is constructed. At each step,
a tree $\tau\in\mathcal{T}$ is picked according to predefined rules.
Using the 2-step procedure of BP, we find the optimal labeling for
$\tau$ using Eq.~(\ref{eq:MAP-cond-tree-energy}). This method includes
the ICM as a special case when the tree is reduced to a single site,
so we call it the tree-based ICM (T-ICM) algorithm, which is presented
in pseudo-code in Algorithm~\ref{alg:Tree-based-ICM}.

\begin{algorithm}
\textbf{Function}: T-ICM()

\textbf{Input}:
\begin{itemize}
\item Graph $\mathcal{G}=(\mathcal{V},\mathcal{E})$.
\item Local cost/energy functions $E_{i}(x_{i},\yb)$ and $E_{ij}(x_{i},x_{j})$.
\item A set of trees $\mathcal{T}$, and an update schedule.
\item Maximum number of iterations $T$.
\item Initial labeling $\xb^{0}$.
\end{itemize}
\textbf{Procedure:}

\textbf{$\quad$For $t=1,2,...,T$} 

\textbf{$\quad$$\quad$}1. \emph{Pick} a conditional tree $\tau$
from the tree set $\mathcal{T}$ according to the update schedule.

\textbf{$\quad$$\quad$}2. \emph{Absorb} neighboring energies according
to Eq.~(\ref{eq:absorbing}).

\textbf{$\quad$$\quad$}3. \emph{Run} $2$-pass BP on the conditional
tree (Section~\ref{sec:Belief-Propagation}): $\xb_{\tau}^{t}\leftarrow BP(\xb_{\tau}^{t-1})$.

\textbf{$\quad$$\quad$}4. \emph{Update} the labeling of the tree:
$\xb_{\tau}^{t}\leftarrow\xb{}_{\tau}^{t-1}$.

\textbf{$\quad$$\quad$}5. \emph{Stop} if $E_{\tau}\left(\x_{\tau}^{t},\xb_{\neighbours(\tau)}^{t},\y\right)$
no longer decreases for all trees $\tau\in\mathcal{T}$.

\textbf{Output}: A local optimal labeling.

\caption{Tree-based iterated conditional mode (T-ICM).\label{alg:Tree-based-ICM}}

\end{algorithm}

\subsubsection{Specification of T-ICM \label{sub:Specification-of-T-ICM}}

\paragraph{Tree set and update schedule.}

For a given graph, there are exponentially many ways to build conditional
trees, and thus defining the tree set is itself a nontrivial task.
However, for grids used in image labeling with height $H$ and width
$W$, we suggest two simple ways:
\begin{itemize}
\item The set of $H$ rows and $W$ columns. The neighborhood of size is
$HS^{W}+WS^{H}$.
\item The set of $2$ alternative rows and $2$ alternative columns (Figs.~\ref{fig:conditional}c,d).
Since alternative rows (or columns) are separated, they can be connected
by \emph{dummy edges} to form a tree (e.g., see Figs.~\ref{fig:conditional}c,d).
A dummy edge has the interacting energy of zero, thus does not affect
search operations on individual components. The neighborhood size
is $4S^{0.5HW}$. 
\end{itemize}
These two sets lead to much more efficient T-ICM compared to the standard
ICM which only covers the neighborhood of size $SHW$ using the same
running time. Once the set has been defined, the update order for
trees can be predefined (e.g., rows from-top-to-bottom then columns
from-left-to-right), or random.

\subsubsection{Properties of T-ICM}

Due to Proposition~\ref{prop:T-ICM-local-min}, at each step of Algorithm~\ref{alg:Tree-based-ICM},
the total energy $E(\xb,\yb)$ will be either reduced or the algorithm
will terminate. Since the model is finite and the energy reduction
is discrete (hence non-vanishing), the algorithm is guaranteed to
reach a local minimum after finite steps.

Although the T-ICM only finds local minima of the energy, we can expect
the quality to be better than the original ICM because each tree covers
many sites. For example, as shown in Figs.~\ref{fig:conditional}a,b,
a tree in the grid can account for half of all the sites, which is
overwhelmingly large compared to a single site used by the ICM. The
number of configurations of the tree $\tau$, or equivalently the
neighborhood size, is $S^{N_{\tau}}$, where $N_{\tau}$ is the number
of sites on the tree $\tau$. The neighborhood size of the ICM, on
the other hands, is just $S$.

For the commonly used $4$-neighbor grid MRF in image labeling, and
the tree set of alternative rows and columns, the BP takes $\mathcal{O}(4HW)$
time to pass messages, each of which cost $3S^{2}$ time to compute.
Thus, the time complexity per iteration of T-ICM is only $S$ times
higher than that of ICM and about the same as that of loop BP (Sec.~\ref{sec:Loopy-BP}).
However, the memory in our case is still $\mathcal{O}\left(2\times\max\{H,W\}\right)$,
which is much smaller than the $\mathcal{O}(4HW)$ memory required
by loopy BP. In addition, each step in T-ICM takes exactly 2 passes,
while the number of iterations of loopy BP for the whole image, if
the method does converge, is unknown and parameter dependent.

\subsection{Tree-based ILS: global search \label{sub:Tree-based-ILS}}

\begin{algorithm}
\textbf{Function:} T-ILS()

\textbf{Input:}
\begin{itemize}
\item Max jump step-size: $\rho_{max}\in(0,100)$.
\item Max number of iterations $T_{outter}$; max number of iterations for
the inner T-ICM $T_{inner}$.
\item Max number of backtracks.
\end{itemize}
\textbf{Procedure:}

$\quad$\emph{Initialize} some labelings: $\tilde{\x}^{0}$.

$\quad$\emph{Find} the first local minimum: $\x^{1}\leftarrow\mbox{T-ICM}\left(\tilde{\x}^{0},T_{inner}\right)$.

$\quad$\emph{Initialize} variables: $n=0$; $\beta=1$.

$\quad$\textbf{For }$t=1,2,...,T_{outter}$

$\quad$$\quad$1. \emph{Jump} to a new place: $\tilde{\x}^{t}\leftarrow\x^{t}$
by randomly resetting $\mathcal{U}\left(0,\rho_{max}\right)\%$ of
labels.

$\quad$$\quad$2. \emph{Find} a local minimum: $\hat{\x}^{t+1}\leftarrow\mbox{T-ICM}\left(\tilde{\x}^{t},T_{inner}\right)$.

$\quad$$\quad$3. \emph{Accept}: $\x^{t+1}\leftarrow\hat{\x}^{t+1}$
with probability of

$\quad$$\quad$$\quad$$\quad$$\quad$$\quad$$a=\min\left\{ 1,\exp\left(-\beta\left\{ E(\hat{\xb}^{t+1},\yb)-E(\x^{t},\yb)\right\} \right)\right\} $

$\quad$$\quad$$\quad$otherwise \emph{backtrack}: $\x^{t+1}\leftarrow\x^{t}$
.

$\quad$$\quad$4. \emph{Adjust} the temperature:

$\quad$$\quad$$\quad$$\quad$$\quad$$\quad$$n\leftarrow n+1$
if $\x^{t+1}=\hat{\x}^{t+1}$; 

$\quad$$\quad$$\quad$$\quad$$\quad$$\quad$$r\leftarrow0.9n/t+0.1\mathbb{I}\left[\x^{t+1}=\hat{\x}^{t+1}\right]$
;

$\quad$$\quad$$\quad$$\quad$$\quad$$\quad$if $r<0.45$ then
$\beta\leftarrow0.8\beta$ else if $r>0.55$ then $\beta\leftarrow\beta/0.8$.

$\quad$$\quad$5. \emph{Stop} if number of backtracks has been reach.

\textbf{Output}: A near global optimal labeling.

\caption{Tree-based iterated local search (T-ILS). \label{alg:Tree-based-ILS}}
\end{algorithm}

As T-ICM is still a local search procedure, inherent drawbacks still
remain: (i) it is sensitive to initialization, and (ii) it can get
stuck in suboptimal solutions. To escape from the local minima, global
search strategies must be employed. We can consider the entire T-ICM
as a single super-move in an \emph{exponentially large neighborhood}. 

\begin{figure}
\begin{centering}
\includegraphics[width=0.8\linewidth]{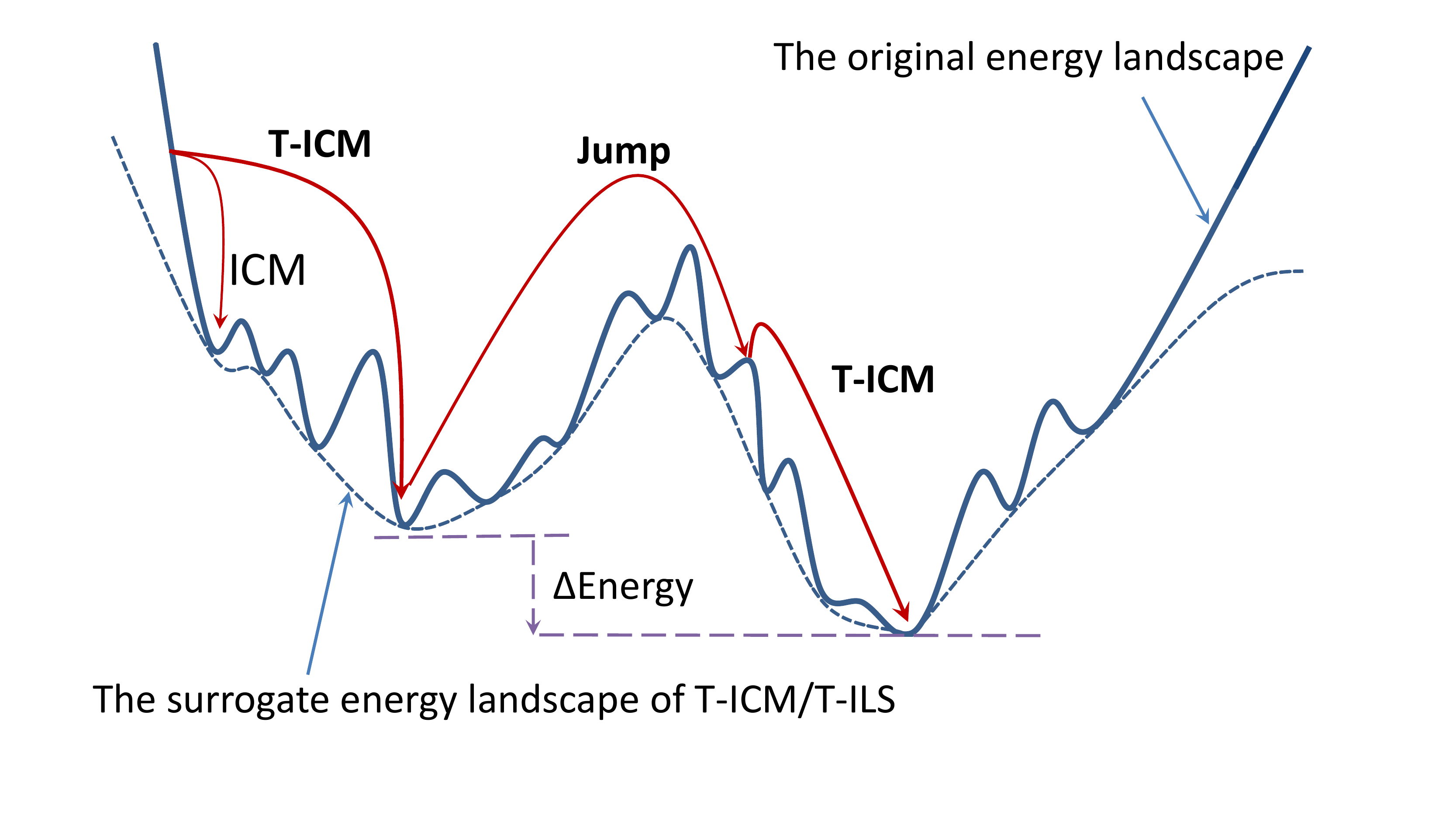}
\par\end{centering}

\caption{Search behavior of T-ICM and T-ILS. The use of T-ICM creates a smoother
energy landscape for T-ILS (the surrogate dotted curve). ICM gets
stuck on the first local minimum it finds, but T-ICM could find a
much better solution by operating on an exponentially large neighborhood.
\label{fig:basin-hop}}
\end{figure}

Since it is not our intention to create a totally new escaping heuristic,
we draw from the rich pool of metaheuristics in the literature and
adapt to the domain of image labeling. In particular, we choose an
effective heuristic, commonly known as iterated local search (ILS)
\cite{Lourenco-et-al03} for escaping from the local minima. ILS advocates
\emph{jumps} from one local minimum to another with the hope that
we can eventually find better local minima after multiple tries. If
a jump fails to lead to a better solution, it can still be accepted
according to an \emph{acceptance} scheme, following the spirit of
simulated annealing (SL). However, we do not decrease the temperature
as in the SL, but rather, the temperature is adjusted so that on average
the acceptance probability is roughly $0.5$. The process is repeated
until the stopping criteria are met. We term the resulting metaheuristic
the \emph{Tree-based Iterated Local Search} (T-ILS) whose pseudo-code
is presented in Algorithm~\ref{alg:Tree-based-ILS}, and behavior
is illustrated in Fig.~\ref{fig:basin-hop}. In what follows we specify
the algorithm in more details.

\subsubsection{Specification of T-ILS \label{sub:Specification-of-T-ILS}}

\paragraph{Jump.}

The jump step-size has to be large enough to successfully escape from
the \emph{basin} that traps the local search. In this study, we design
a simple jump by randomly changing labels of $\rho\%$ of sites. The
step-size $\rho$ is drawn randomly in the range $(0,\rho_{max})$,
i.e., $\rho\sim\mathcal{U}\left(0,\rho_{max}\right)$, where $\rho_{max}$
is an user-specified parameter.

\paragraph{Acceptance.}

After a jump, the local search is invoked, followed by an acceptance
decision to accept or reject the jump. Convergence guaranteed acceptance
criteria such as those used in simulated annealing can be used, but
it is likely to be slow. We consider the following acceptance probability:

\[
a=\min\left\{ 1,\exp(-\beta\Delta E)\right\} 
\]
where $\Delta E=E(\hat{\xb}^{t+1},\yb)-E(\x^{t},\yb)$ is the change
in energy between two consecutive minima, and $\beta>0$ is the adjustable
``inverse temperature''. Large $\beta$ lowers the acceptance rate
but small $\beta$ increases the rate. This fact will be used to adjust
the acceptance rate, as detailed below.

\paragraph{Adjusting inverse temperature.}

We wish to maintain an average acceptance probability of $0.5$, following
the success of \cite{wales1997global}. However, unlike the work in
\cite{wales1997global}, we do not change the step-size, but rather
adjusting the inverse temperature. The estimation of acceptance rate
is $r\leftarrow r/t$, where $n$ is the total number of accepted
jumps up to step $t$. To introduce short-term effect, we use the
last event: 
\[
r\leftarrow0.9r+0.1\mathbb{I}\left[\x^{t+1}=\hat{\x}^{t+1}\right]
\]
If the acceptance rate is within the range $[0.45,0.55]$ we do nothing.
A lower rate will lead to decrease of the inverse temperature: $\beta\leftarrow0.8\beta$,
and a higher rate will lead to increase: $\beta\leftarrow\beta/0.8$.

\subsubsection{Properties of T-ILS}

Fig.~\ref{fig:basin-hop} illustrates the behavior of the T-ILS.
Through the T-ICM component, the energy landscape is smoothed out,
helping the T-ICM to locate good local minima. When the jump is not
large, the search trajectory can be tracked to avoid self-crossing
walks. If the jump is far enough (with large $\rho_{max}$), the resulting
algorithm will behave like the well-known multistart procedures. 

When the inverse temperature $\beta$ is set to $0$, the acceptance
become deterministic, that is, we only accept the jump if it improves
the current solution. In other words, the T-ILS becomes a greedy algorithm.
Alternatively, when $\beta$ is set to a very large number and no
adjustment is made, we would accept all the jumps, allowing memoryless
foraging behavior.

\section{Experiments\label{sec:exp}}

In this section, we evaluate our proposed algorithms on a simulated
\emph{Ising model} and two standard vision labeling problems: \emph{stereo
correspondence }and\emph{ image denoising}. In all settings, we employ
MRFs with grid-structure (e.g. each inner pixel is connected to exactly
4 nearby pixels). Trees are composed of rows and columns as specified
in Sec.~\ref{sub:Specification-of-T-ICM}. Unless specified otherwise,
the initial labeling is randomly assigned. Max step-size is $\rho_{max}=10\%$
(Sec.~\ref{sub:Specification-of-T-ILS}). For T-ILS, the inner loop
has $T_{inner}=1$, i.e., the full local minima may not be reached
by the T-ICM, as does not seem to hurt the final performance.

\subsection{Simulated Ising model \label{sub:Simulated-Ising-model}}

In this subsection, we validate the robustness of our proposed algorithms
on Ising models, which have wide applications in magnetism, lattice
gases, and neuroscience \cite{mccoy1973two}. Within the MRF literature,
Ising lattice is often used as a benchmark to test inference algorithms
(e.g., see \cite{Wainwright-et-al05TIT}). Following \cite{Wainwright-et-al05TIT},
we simulate a 500$\times$500 grid Ising model where labels are binary
spin orientations (up or down): $x_{i}\in\pm1$, and local energy
functions are: $E_{i}(x_{i})=\theta_{i}x_{i}$; $E_{ij}(x_{i},x_{j})=\theta_{ij}x_{i}x_{j}$.
The parameter $\theta_{i}$ specifies the influence of external field
on the spin orientation and $\theta_{ij}$ specifies the interaction
strength and direction (impulsive or repulsive) between sites. For
this experiment, the parameters $\left\{ \theta_{i},\theta_{ij}\right\} $
are set as follows 
\begin{eqnarray*}
\theta_{i},\theta_{ij} & \sim & \mathcal{U}(-1,1)
\end{eqnarray*}
where $\mathcal{U}(-1,1)$ denotes the uniform distribution in the
range $(-1,1)$, and $\lambda>0$ specifies the interaction strength.
When $\lambda$ is small, the interaction is weak, and thus the external
field has more effect on the spin arrangement. However, when $\lambda$
is large, the spin arrangement depends more on the interacting nature.
A stable arrangement in nature would be of the minimum energy. 

\begin{figure}
\begin{centering}
\begin{tabular}{cc}
\includegraphics[width=0.45\linewidth]{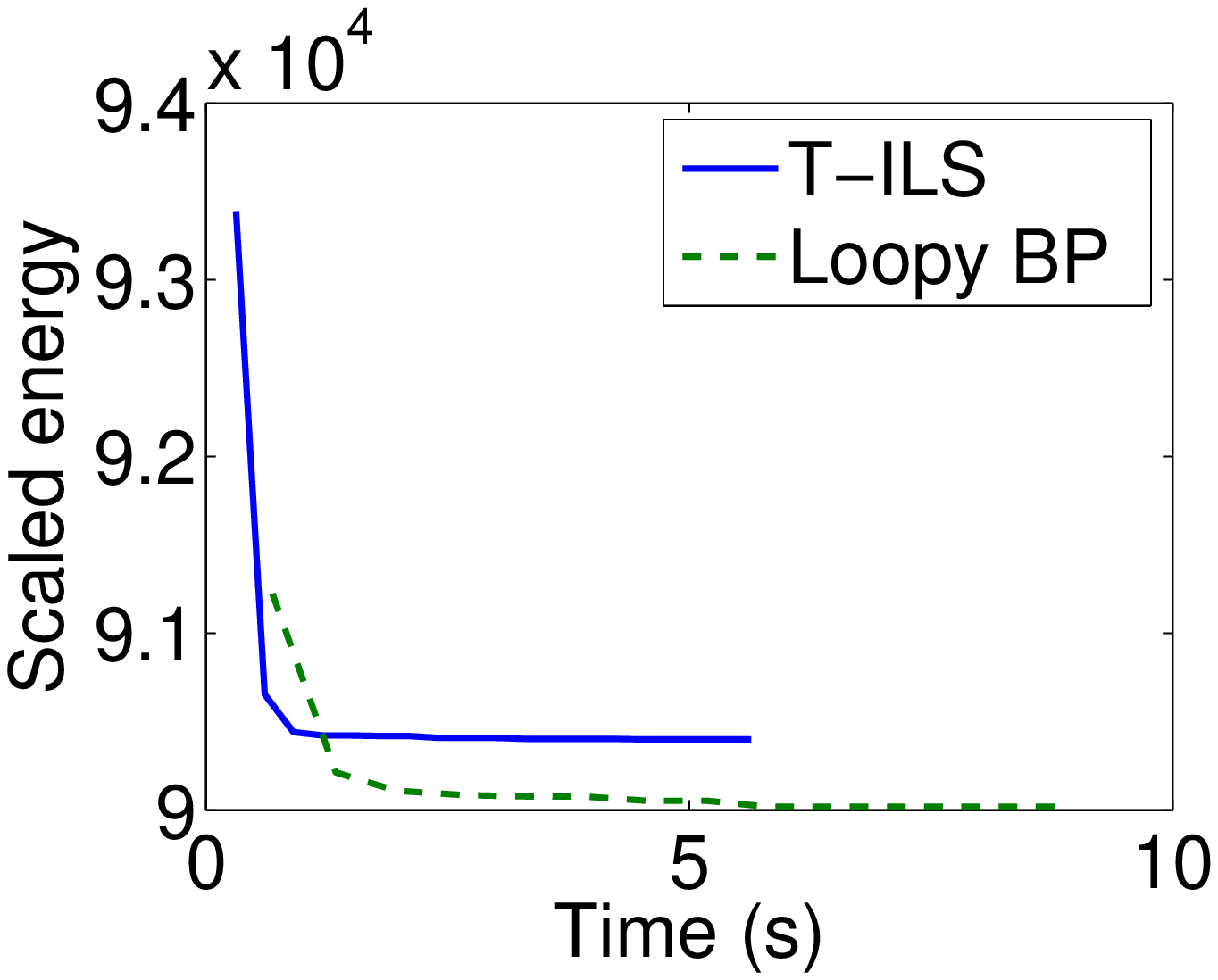}  & \includegraphics[width=0.45\linewidth]{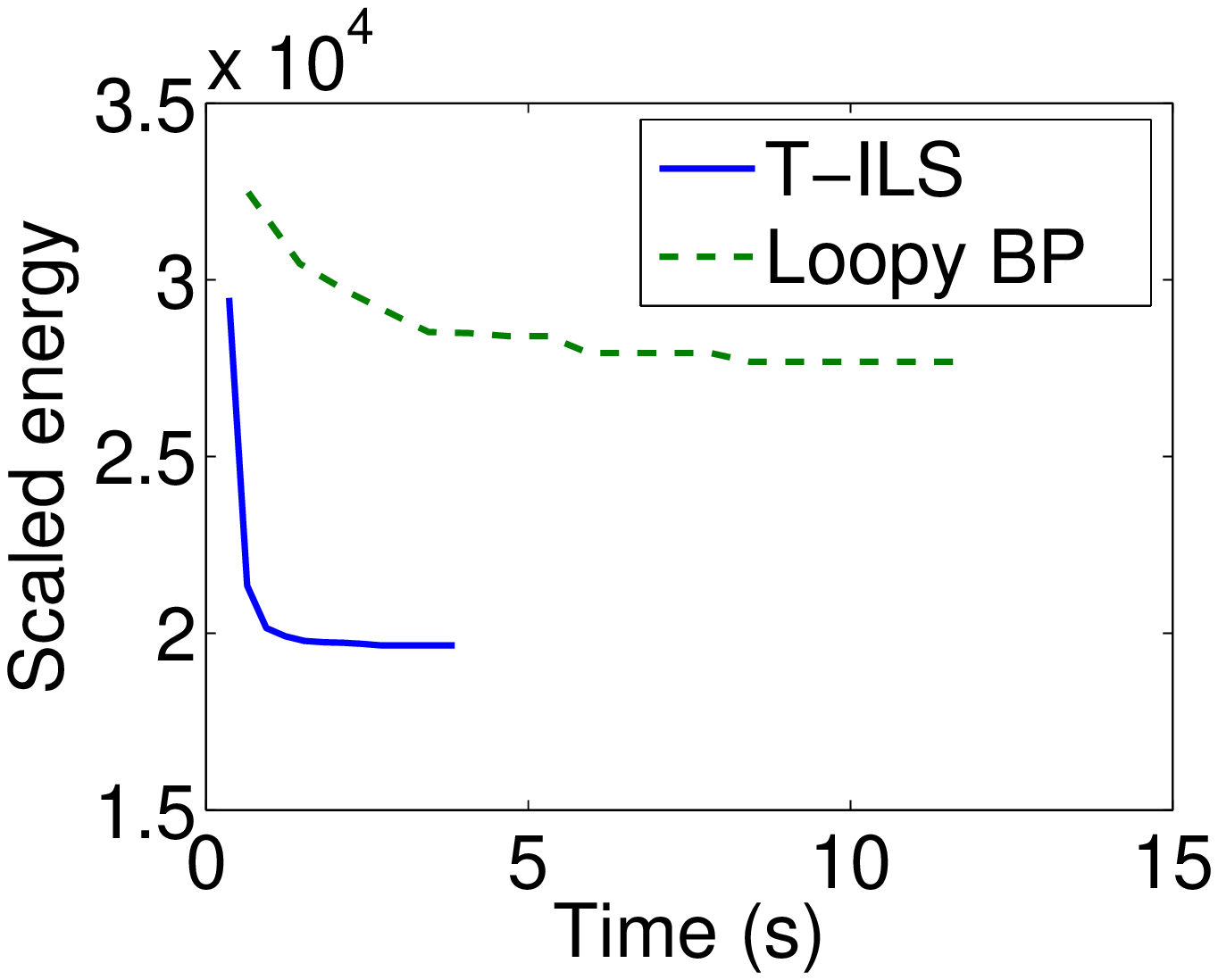} \tabularnewline
\end{tabular}
\par\end{centering}

\caption{Performance of T-ILS and loopy BP algorithm in minimizing Ising energy
with $\lambda=0.5$ (left) and $\lambda=1.0$ (right). \label{fig:Ising}}
\end{figure}

The result of minimizing energy is shown in Fig.~\ref{fig:Ising}.
When the interaction is weak (e.g. $\lambda=0.5$), the loopy BP performs
well, but when the interaction is strong (e.g. $\lambda=1.0$), the
T-ILS has a clear advantage. Thus T-ILS is more robust since it is
less sensitive to $\lambda$.

\subsection{Stereo correspondence \label{sub:Stereo-correspondence}}

\begin{figure}
\begin{centering}
\begin{tabular}{ccc}
\includegraphics[width=0.297\linewidth]{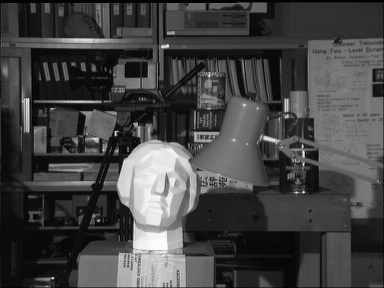} & \includegraphics[width=0.297\linewidth]{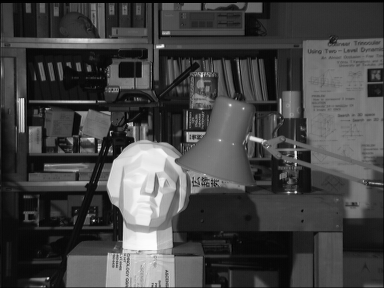} & \includegraphics[width=0.297\linewidth]{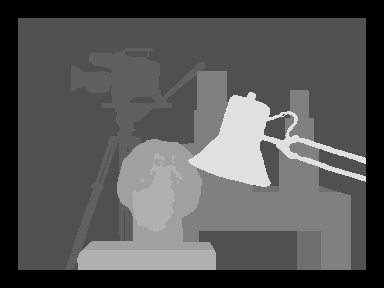}\tabularnewline
 \includegraphics[width=0.3\linewidth]{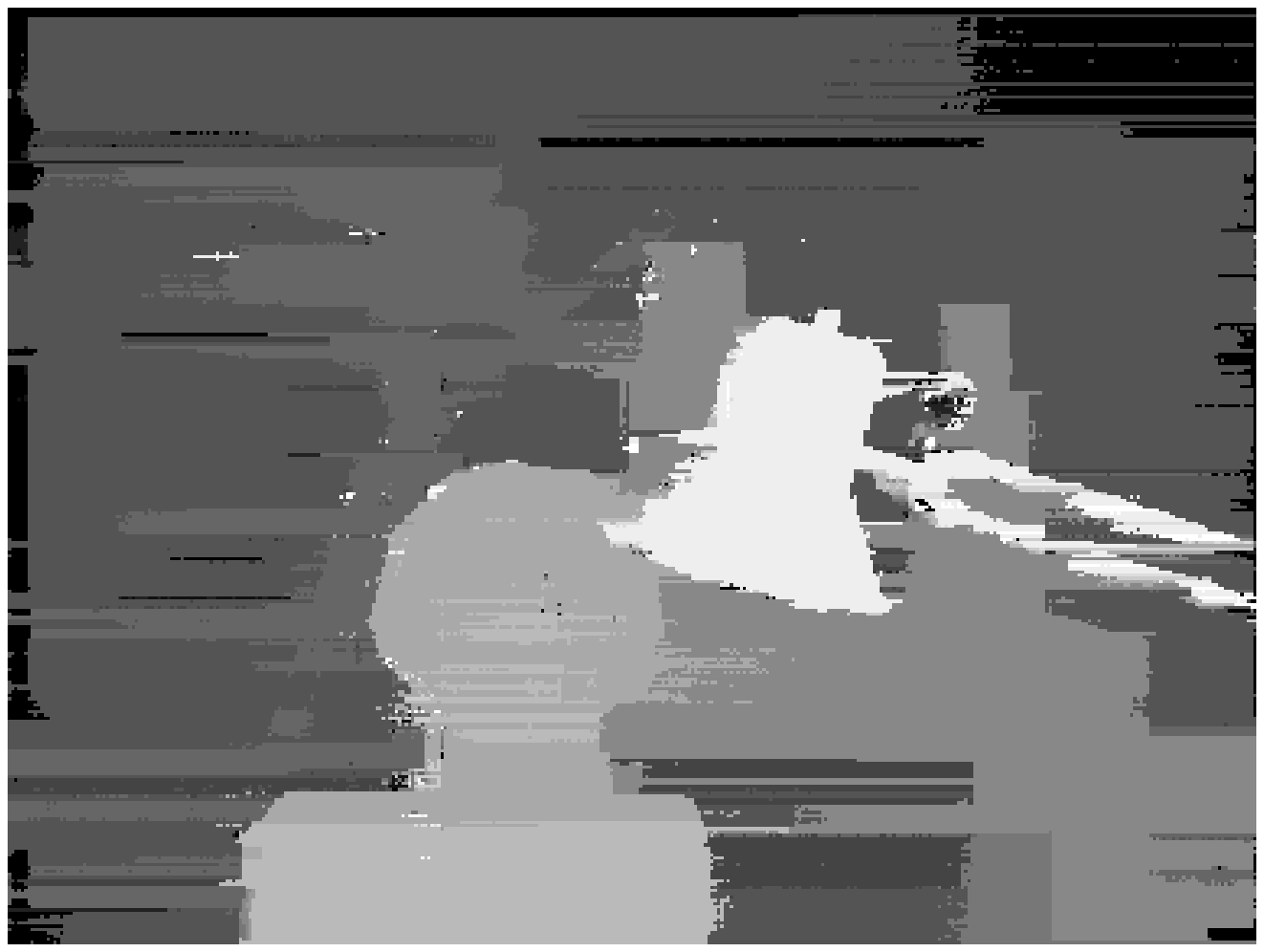} & \includegraphics[width=0.3\linewidth]{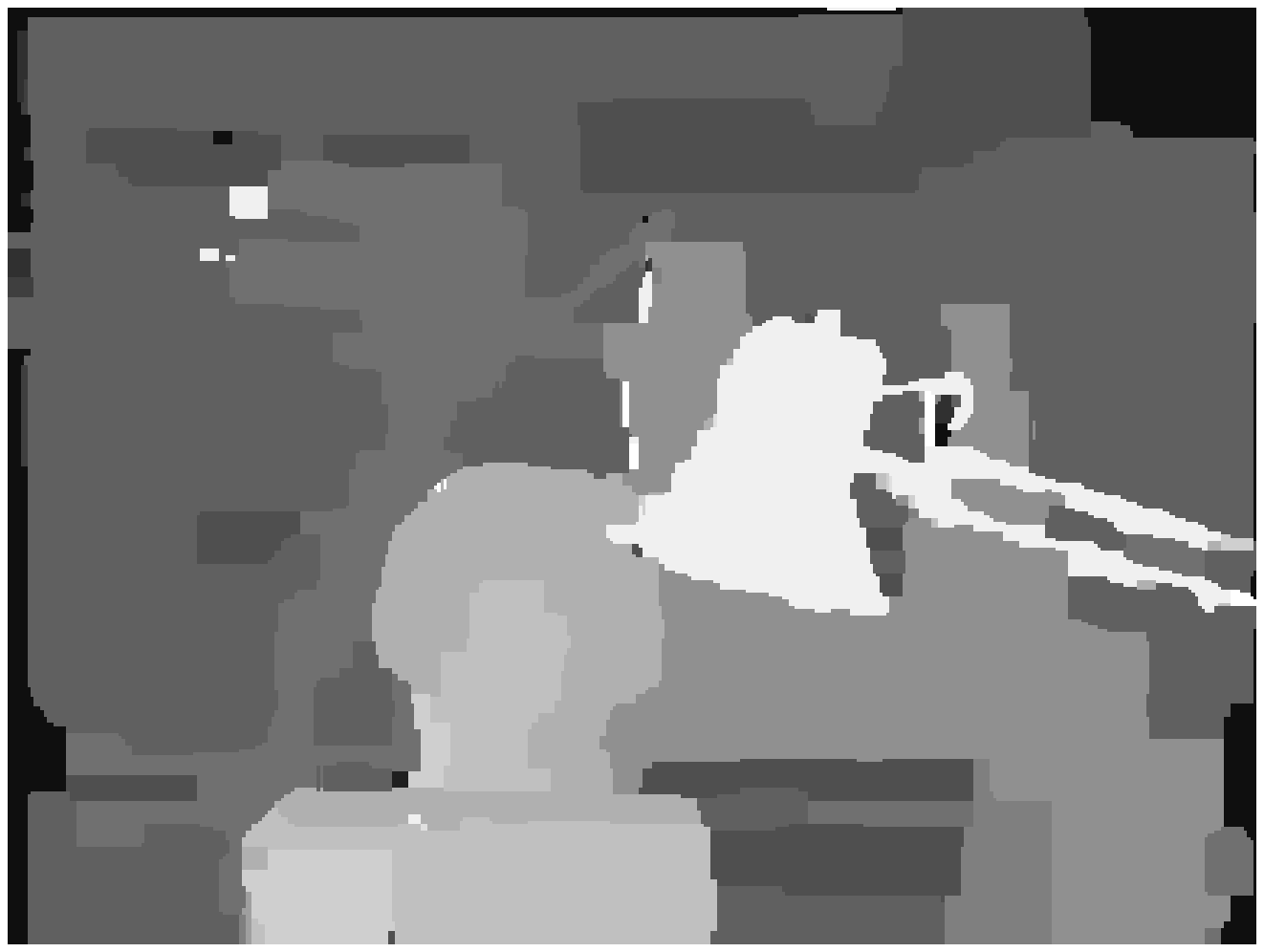} & \includegraphics[width=0.3\linewidth]{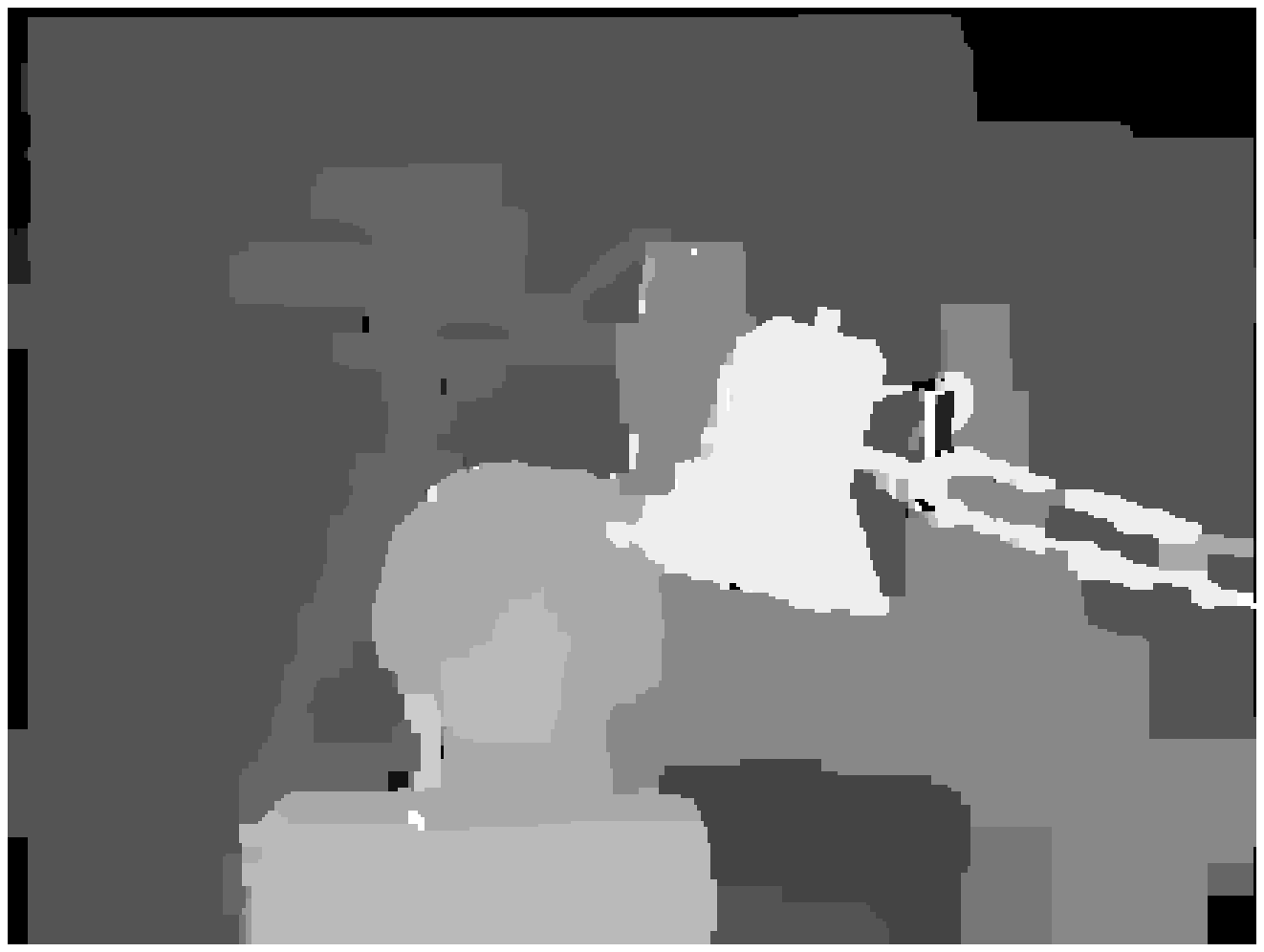}\tabularnewline
\end{tabular}
\par\end{centering}

\caption{Stereo results on the Tsukuba dataset. (Top-left): left image, (top-middle):
right image, (top-right): groundtruth; (bottom-left): scan-line, (bottom-middle):
loop BP, and (bottom-right): T-ILS (initialized from scan-line). \label{fig:tsukuba}}
\end{figure}

\begin{figure}
\begin{centering}
\begin{tabular}{ccc}
\includegraphics[width=0.297\linewidth]{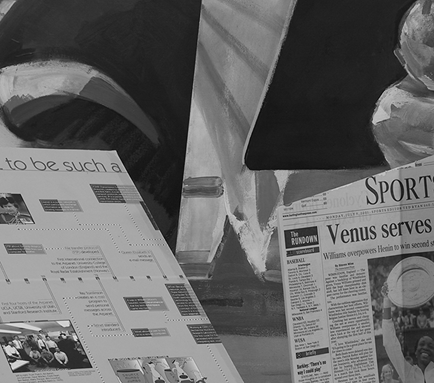} & \includegraphics[width=0.297\linewidth]{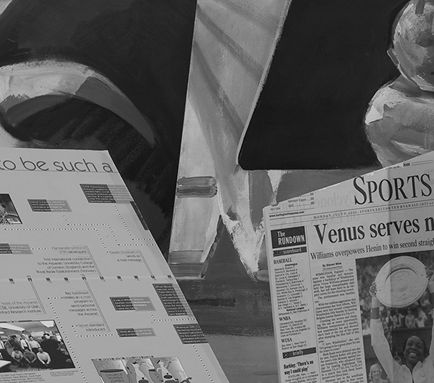} & \includegraphics[width=0.297\linewidth]{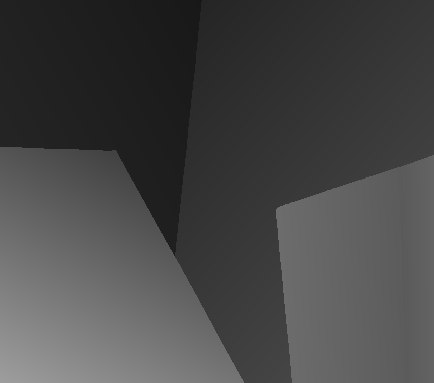}\tabularnewline
 \includegraphics[width=0.3\linewidth]{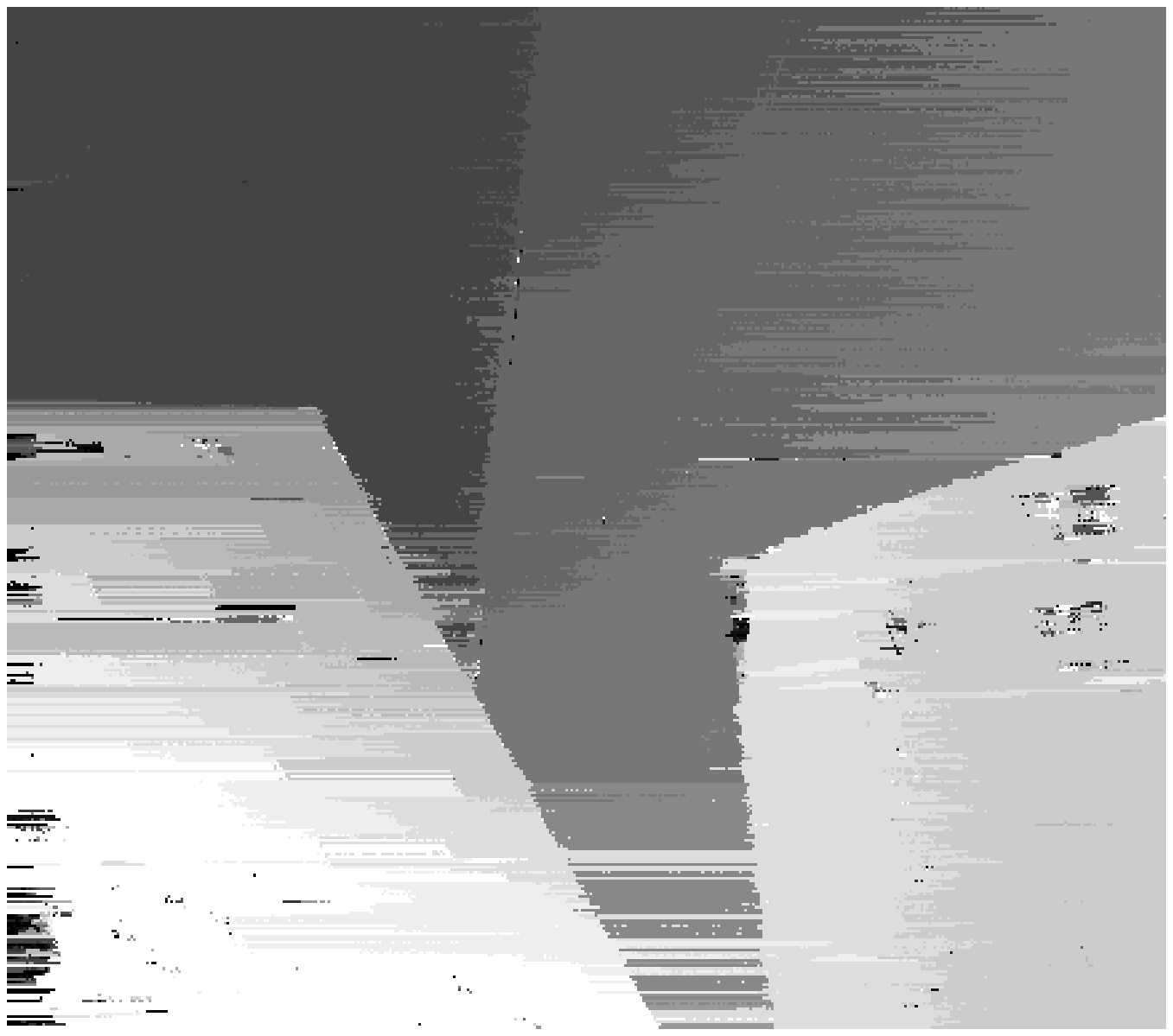} & \includegraphics[width=0.3\linewidth]{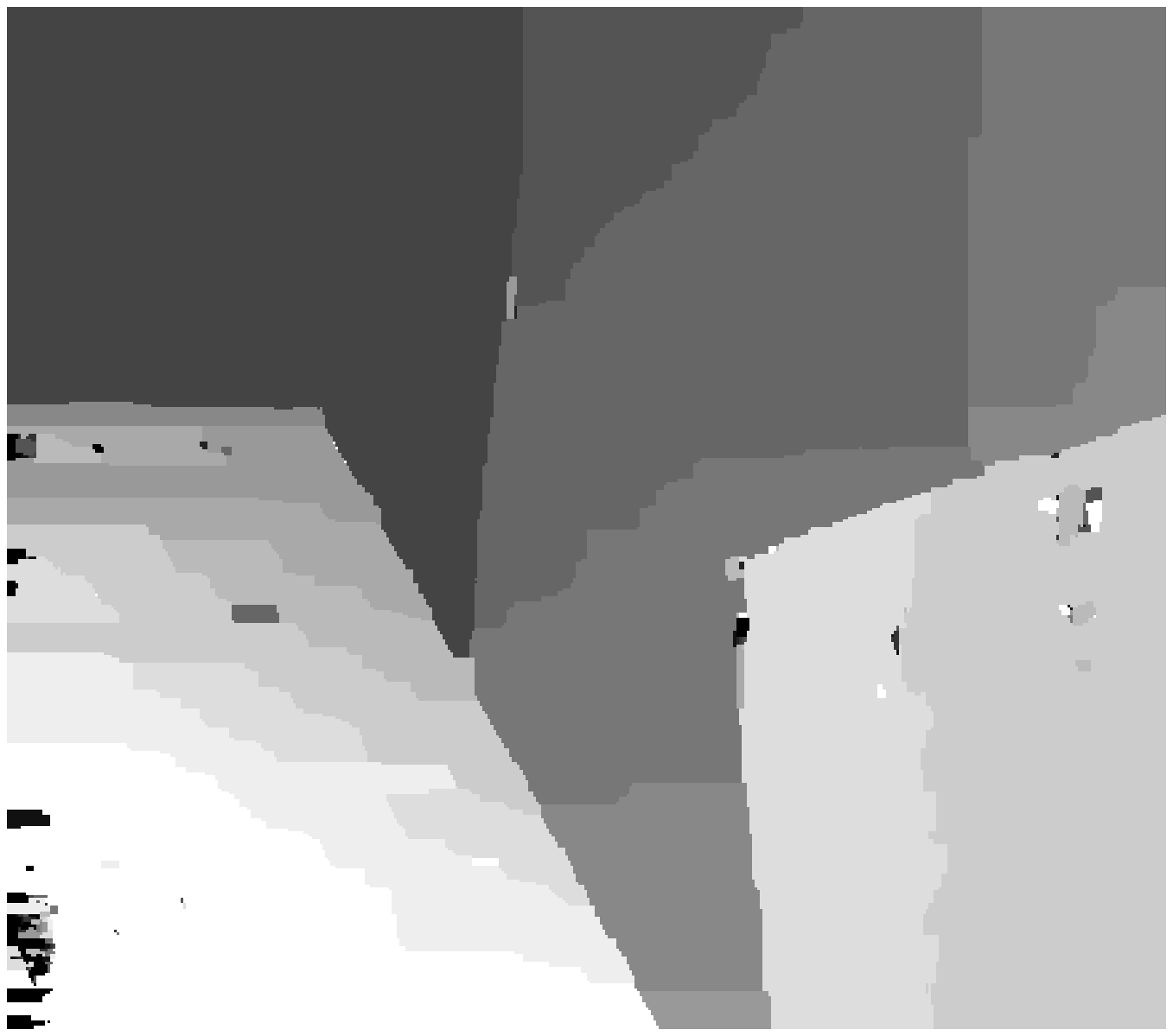} & \includegraphics[width=0.3\linewidth]{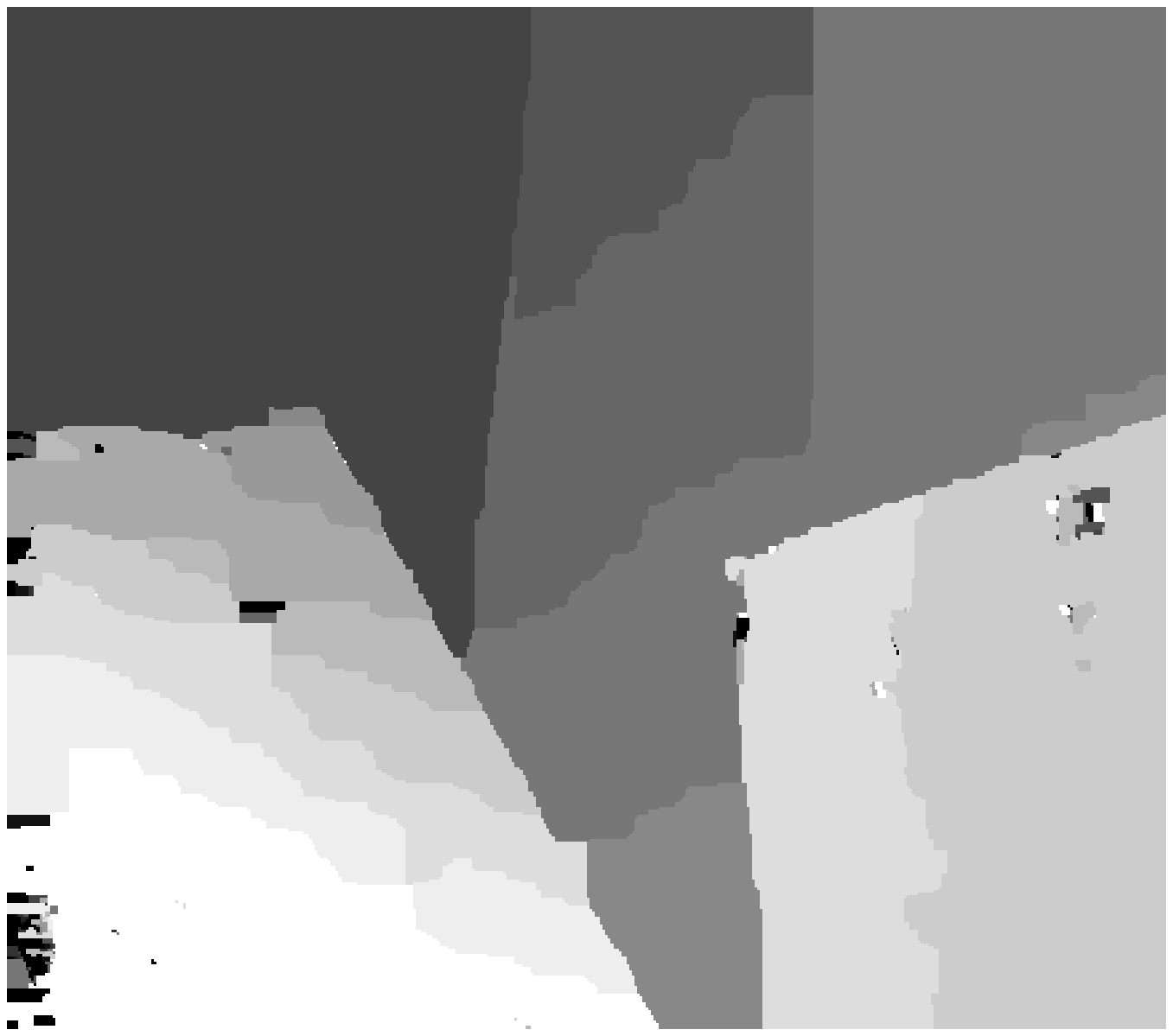}\tabularnewline
\end{tabular}
\par\end{centering}

\caption{Stereo results on the Venus dataset. (Top-left): left image, (top-middle):
right image, (top-right): groundtruth; (bottom-left): scan-line, (bottom-middle):
loop BP, and (bottom-right): T-ILS (initialized from scan-line). \label{fig:venus}}
\end{figure}

The problem in stereo correspondence is to estimate the depth of the
field (DoF) given two or more 2D images of the same scene taken from
two or more cameras arranged horizontally. This is used in 3D reconstruction
of a scene using standard 2D cameras. The problem is often translated
into estimating \emph{disparit}y between images -- how much two images
differ and this reflects the depth at any pixel locations. For simplicity,
we only investigate the two-cameras setting. In the MRF-based stereo
framework, a configuration of $\xb\in\mathbb{N}^{W\times H}$ realizes
the \emph{disparity map}, which in this case is represented by a grid
network. The disparity set (or the label set) is often predefined.
For example, in the two standard datasets%
\footnote{Available at: http://vision.middlebury.edu/stereo/%
} used in this experiment, the Tsukuba has $16$ labels (Fig.~\ref{fig:tsukuba}),
and the Venus has $20$ (Fig.~\ref{fig:venus}). 

The singleton energy $E_{i}(x_{i},\y)$ at each pixel location usually
measures the dissimilarity of pixel intensity between the left/right
images, and the interaction energy $E_{ij}(x_{i},x_{j})$ ensures
the smoothness of the disparity map. In this set of experiments, we
use the simple linear Potts cost model as often used in testing stereo
correspondence algorithms \cite{Scharstein-Szeliski02}. Let $\y=(I^{l},I^{r})$
where $I^{l}$ and $I^{r}$ are intensities of the left and right
images respectively, and $i=(i_{X},i_{Y})$ where $i_{X}$ and $i_{Y}$
are horizontal and vertical coordinates of the pixel $i$. The local
energies are defined as \cite{Scharstein-Szeliski02}: 
\begin{eqnarray*}
E_{i}(x_{i},\y) & = & \Delta I(i,x_{i})\\
E_{ij}(x_{i},x_{j}) & = & \lambda\times\mathbb{I}[x_{i}\ne x_{j}]
\end{eqnarray*}
where $\mathbb{I}[\cdot]$ is the indicator function, $\lambda>0$
is the smoothness parameter, and$\Delta I(i,x_{i})=\left|I^{l}(i_{X},i_{Y})-I^{r}(i_{X}-x_{i},i_{Y})\right|$
is the difference in pixel intensity in two images when pixel positions
are $x_{i}$ pixels apart in the horizontal dimension. Due to the
optical properties of two nearby cameras, a small $x_{i}$ would result
in large $\Delta I(i,x_{i})$ if the true DoF is small. Thus by minimizing
the singleton energy with respect to $x_{i}$, a small DoF would leads
to stronger reduction of $x_{i}$ than a large DoF. For this set of
experiments, we choose $\lambda=20$ following \cite{Scharstein-Szeliski02}.
Our implementation is based on the software framework of \cite{szeliski2007comparative}%
\footnote{The C++ code is available at http://vision.middlebury.edu/MRF/%
}.

There is a wide range of techniques available for stereo estimation,
and loopy BP is one of the winning methods \cite{Scharstein-Szeliski02}\cite{szeliski2007comparative}.
Fast methods like \emph{scan-line }(SL) optimization are still widely
used for real-time implementation. The scan-line is equivalent to
taking independent 1D rows of the MRF and running the chain BP. Since
the SL does not admit the original 2D structure, we need to adapt
the singleton energy as: $\bar{E}_{i}(x_{i},\y)=\nu E_{i}(x_{i},y)$
where $\nu\in[0,1]$ to account for the lack of inter-row interactions.

\begin{table}[htb]
\begin{centering}
\begin{tabular}{|l|rr|}
\hline 
Method  & Tsukuba & Venus\tabularnewline
\hline 
SL($\nu=1.0$)  & 814,121 & 1,362,067\tabularnewline
SL($\nu=0.4$)  & 658,946 & 1,198,324\tabularnewline
Random$\rightarrow$T-ICM($T=1)$  & 739,370 & 1,048,587\tabularnewline
SL($\nu=0.4$)$\rightarrow$T-ICM($T=1)$  & 427,860 & 669,973\tabularnewline
Loopy BP($T=1,000$) & 413,269  & 640,385 \tabularnewline
SL($\nu=0.4$)$\rightarrow$T-ILS($T_{outter}=1,000$) & \textbf{403,129}  & \textbf{635,305} \tabularnewline
\hline 
\end{tabular}
\par\end{centering}

\caption{Stereo energy found by algorithms. SL=Scan-line.\label{tab:stereo-energy-time} }
\end{table}

\begin{figure}
\begin{centering}
\begin{tabular}{cc}
\includegraphics[width=0.48\textwidth]{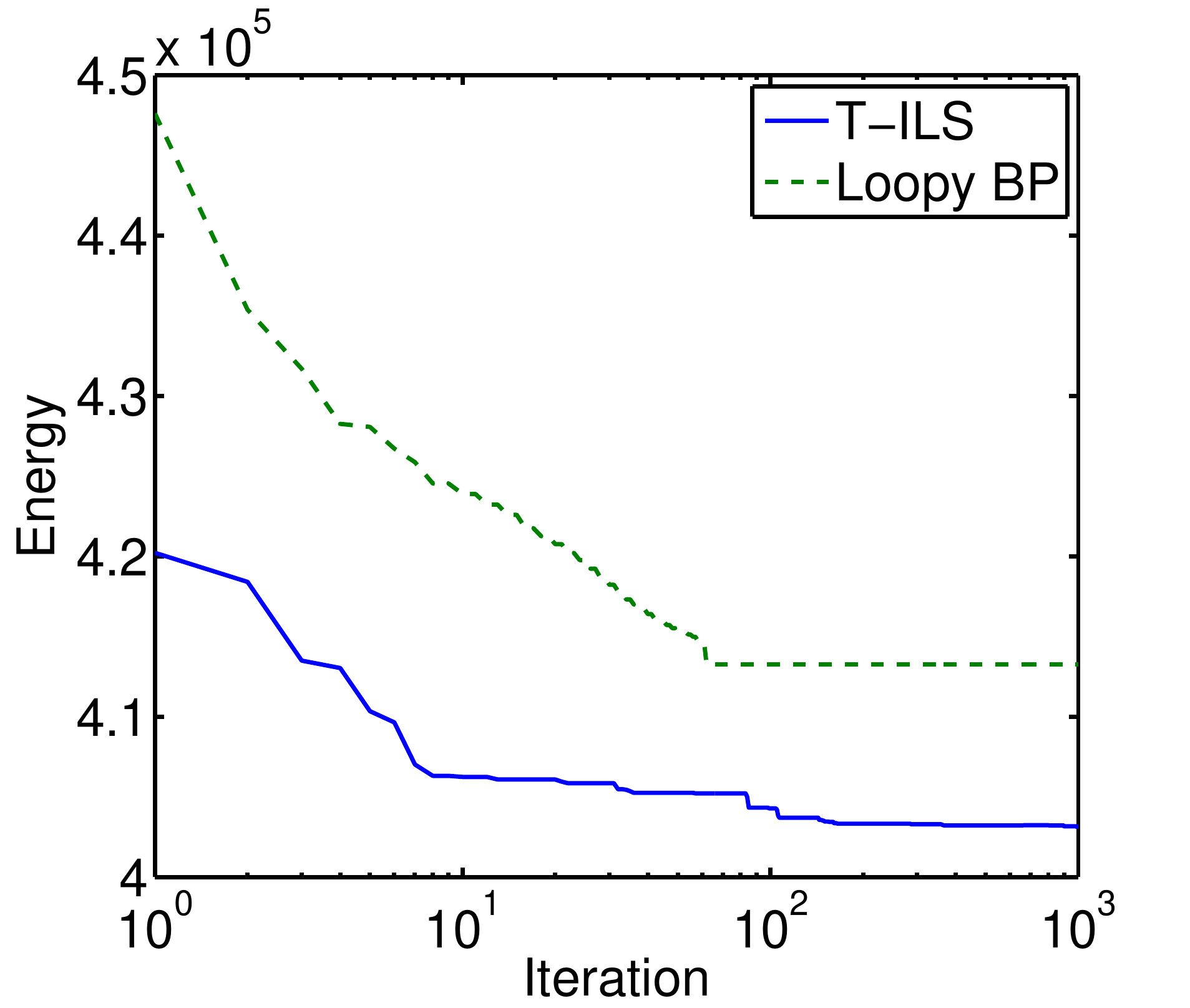} & \includegraphics[width=0.48\textwidth]{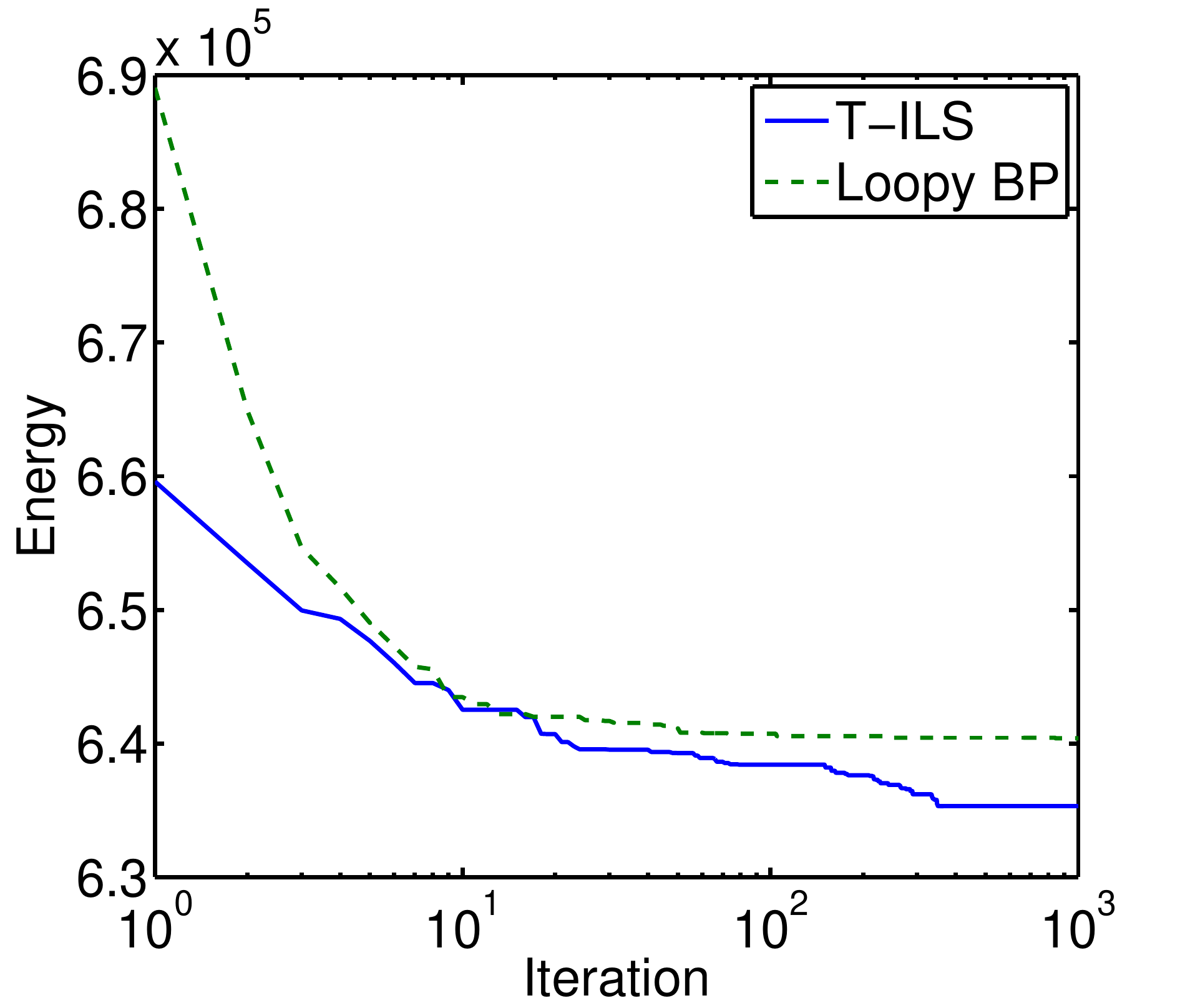}\tabularnewline
Tsukuba (Fig\@.~\ref{fig:tsukuba}) & Venus (Fig\@.~\ref{fig:venus})\tabularnewline
\end{tabular}
\par\end{centering}

\caption{Stereo energy minimization by Loopy BP (dashed line) and T-ILS (line)
on Tsukuba (left) and Venus (right) images. \label{fig:Stereo-energy-minimization}}

\end{figure}

Table~\ref{tab:stereo-energy-time} shows the effect of changing
from $\nu=1.0$ to $\nu=0.4$ in term of reducing 2D energy and error.
The result, however, has the inherent horizontal `streaking' effect
since no 2D constraints are ensured (Figs.~\ref{fig:tsukuba},\ref{fig:venus},
bottom-left). The randomly initialized T-ICM with one iteration ($T=1$
in Algorithm~\ref{alg:Tree-based-ICM}) performs comparably with
the best of SL ($\nu=0.4$). The performance of T-ICM improves significantly
by initializing from the result of SL. T-ILS initialized from SL finds
better energy than the loopy BP given the same number of iterations,
as shown in Fig.~\ref{fig:Stereo-energy-minimization}.

\subsection{Image denoising \label{sub:Image-denoising}}

\begin{figure}
\begin{centering}
\begin{tabular}{>{\centering}p{0.2\textwidth}>{\centering}p{0.15\textwidth}>{\centering}p{0.15\textwidth}c}
\multicolumn{3}{c}{\includegraphics[height=0.28\textwidth]{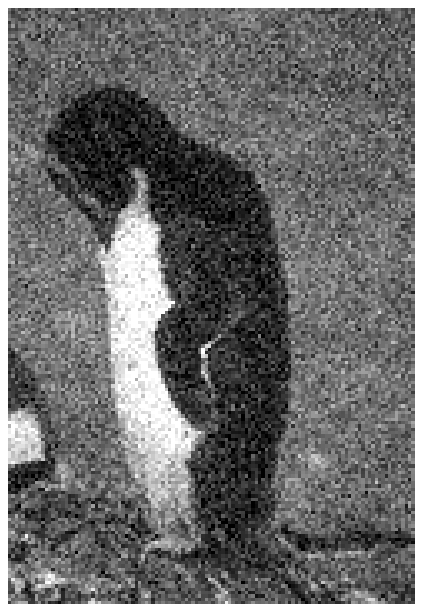} \includegraphics[height=0.28\textwidth]{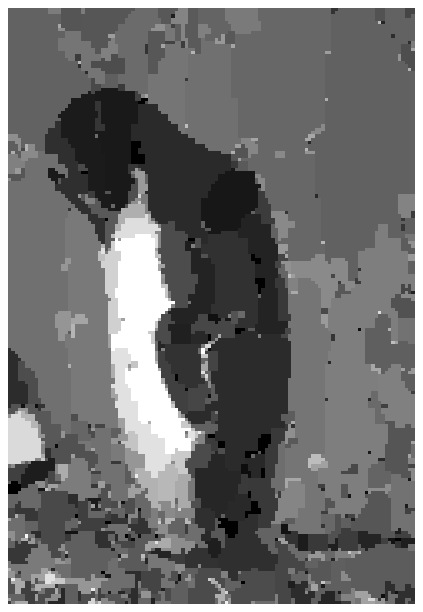}~\includegraphics[height=0.28\textwidth]{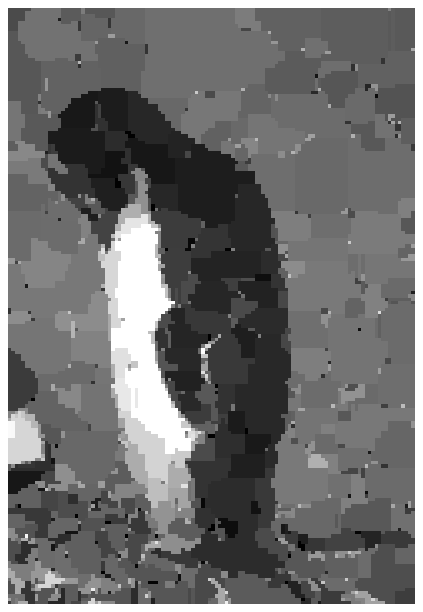}} & \includegraphics[height=0.28\textwidth]{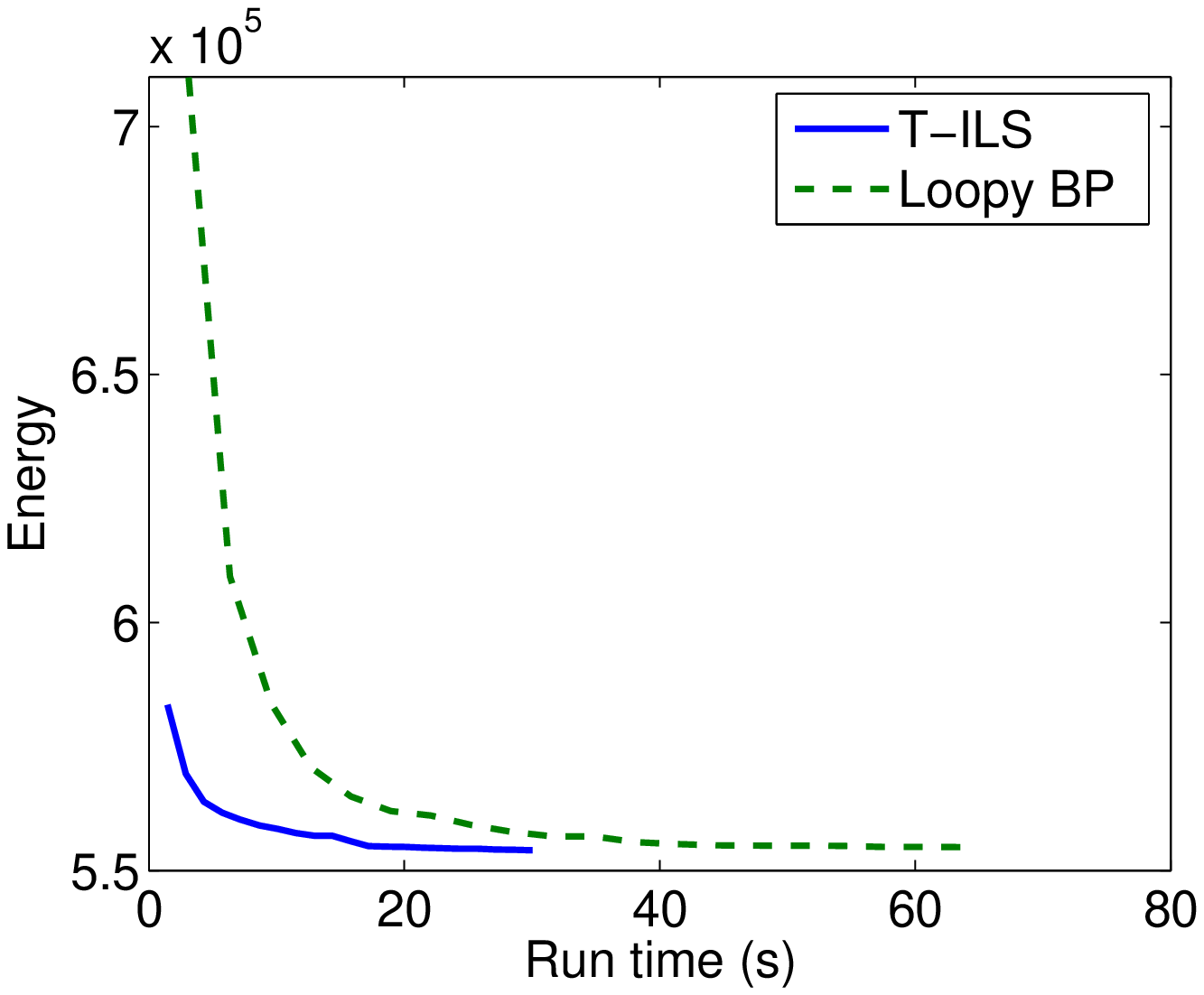}\tabularnewline
(a) & (b) & (c) & (d)\tabularnewline
\end{tabular}
\par\end{centering}

\caption{Penguin images: (a) noisy, (b) restored with T-ILS, (b) restored with
loopy BP; (d) running time. Algorithms stop after $5$ unsuccessful
iterations. \label{fig:penguin-restore}}
\end{figure}

In image denoising, we are given an image corrupted by noise, and
the task is to reconstruct the original image. For this set of experiments
we use the $122\times179$ noisy gray Penguin image%
\footnote{Available at: http://vision.middlebury.edu/MRF/%
} (Fig.~\ref{fig:penguin-restore}). The labels of the MRF correspond
to $S=256$ intensity levels ($8$ bits depth). Similar to the stereo
correspondence problem, we use a simple truncated Potts model for
the energy as follows 
\begin{eqnarray*}
E_{i}(x_{i},\y) & = & \min\left\{ |x_{i}-y_{i}|,\tau\right\} \\
E_{ij}(x_{i},x_{j}) & = & \lambda\times\delta[x_{i}\ne x_{j}]
\end{eqnarray*}
where $\tau=100$ prevents the effect of extreme noise, and $\lambda=25$
is smoothness parameter, following \cite{szeliski2007comparative}.
In addition, the optimized loopy BP for Potts models from \cite{Felzenszwalb-Huttenlocher-IJCV06}
is used. Figs~\ref{fig:penguin-restore}(b,c) demonstrates that T-ILS
runs faster than the optimize loopy BP, yielding lower energy and
smoother restoration.

\section{Discussion \label{sec:conclusion} }

We have proposed a fast method for inference in Markov random fields
by exploiting tree structures embedded in the network. We proposed
a strong local search operator (T-ICM) based on Belief-Propagation
and a global stochastic search operator T-ILS based on the iterated
local search framework. We have shown in both simulation and two real-world
image analysis tasks (stereo correspondence and image denoising) that
the T-ILS is competitive against state-of-the-art algorithms. We have
demonstrated that by exploiting the structure of the domains, we can
derive strong local search operators which can be exploited in a metaheuristic
strategy such as ILS.

\subsection{Future work}

The line of current work could be extended in several directions.
First, we could adapt the T-ILS for certain cost functions. Currently,
the T-ILS is designed as a generic optimization method, making no
assumptions about the the nature of optimal solution. In contrast,
label maps in vision are often smooth almost everywhere except for
sharp boundaries. Second, the MRFs may offer a more informative way
to perform the jump steps, e.g., by relaxing the messages in the local
edges or by keeping track of the hopping trajectories. Third, we have
limited the T-ILS to uniform distribution of step-sizes, but it needs
not be the case. One useful heuristic is L\'{e}vy flights \cite{tran2004global,yang2009cuckoo}
in which the step-size $\rho$ is drawn from the power-law distribution:
$\rho^{-\alpha}$ for $\alpha>0$. This distribution allows the step-size
to usually stay within a certain narrow range but also allows occasional
big jump (which might behave like a total restart). Alternatively,
we can use Beta distribution due to its richness in distribution shape
and the boundedness of $\rho$. Fourth, since a MRF typically contain
many separated conditional trees, it could be interesting to explore
the parallel jump strategies.

As our development is based on the recognition that we can exploit
conditional trees in MRFs to obtain fast strong local search. The
metaheuristics thus need not be limited to ILS. In fact, conditional
trees open up a new direction of research in the MAP assignment by
investigating the use of other metaheuristics. For example, we can
use genetic algorithms in conjunction with the conditional trees as
follows. Each individual in the GA population can be represented by
a string of $N$ characters, each of which has one of $S$ values
in the label alphabet. For each individual, we run the 2-pass BP to
obtain a strong local solution. Then the crossover operator can be
applied character-wise on a selected subset to generate a new population.

Finally, although we have limited ourselves to applications in image
analysis, the proposed algorithm is generic to any problems where
MRFs are applicable. 

\appendix

\section{Appendix}

\subsection{Distribution over conditional trees \label{sub:Distribution-over-conditional}}

We provide the derivation of Eq.~(\ref{eq:cond-tree-energy}) from
the probabilistic argument. Recall that $\neighbours(\tau)$ is the
Markov blanket of the tree $\tau$, that is, the set of sites connecting
to $\tau$. Due to the Markov property

\[
P\left(\x_{\tau}\mid\xb_{\neighbours(\tau)},\y\right)=P\left(\x_{\tau}\mid\xb_{\neg\tau},\y\right)
\]
In other words, we need only to worry about the interaction within
the tree, and between the tree sites with its Markov blanket:

\[
P\left(\x_{\tau}\mid\xb_{\neighbours(\tau)},\y\right)\propto\exp\left\{ -E_{\tau}\left(\x_{\tau},\xb_{\neighbours(\tau)},\y\right)\right\} 
\]
where
\begin{eqnarray}
E_{\tau}\left(\x_{\tau},\xb_{\neighbours(\tau)},\y\right) & = & \sum_{i\in\tau}E_{i}(x_{i},\yb)+\sum_{(i,j)\in\mathcal{E}\mid i,j\in\tau}E_{ij}(x_{i},x_{j})+\sum_{(i,j)\in\mathcal{E}\mid i\in\tau,j\in\neighbours(\tau)}E_{ij}(x_{i},x_{j})\label{eq:conditional-tree-energy}
\end{eqnarray}

Eq.~(\ref{eq:cond-tree-energy}) can be derived from the energy above
by letting: 
\begin{equation}
E_{i}^{*}(x_{i},\yb)=E_{i}(x_{i},\yb)+\sum_{(i,j)\in\mathcal{E},j\in\neighbours(\tau)}E_{ij}(x_{i},x_{j})\label{eq:absorbing-1-1}
\end{equation}
Thus finding the most probable labeling of the tree $\tau$ conditioned
on its neighborhood is equivalent to minimizing the conditional energy
in Eq.~(\ref{eq:MAP-cond-tree-energy}): 
\begin{equation}
\hat{\x}_{\tau}=\arg\max_{\x_{\tau}}P\left(\x_{\tau}\mid\xb_{\neighbours(\tau)},\y\right)\label{eq:MAP-cond-tree-prob}
\end{equation}
The equivalence can also be seen intuitively by considering the tree
$\tau$ as a mega-site, so the update in Eq.~(\ref{eq:MAP-cond-tree-energy})
is analogous to that in Eq.~(\ref{eq:ICM-energy}).

\subsection{Proof of Proposition~\ref{prop:T-ICM-local-min} \label{sub:Proof-of-T-ICM-local-opt}}

Recall that the energy can be decomposed into singleton and pairwise
local energies (see Eq.~(\ref{eq:energy-decompose}))
\begin{eqnarray*}
E(\x_{\tau},\x_{\neg\tau},\y) & = & \sum_{i\in\tau}E_{i}(x_{i},\y)+\sum_{i\notin\tau}E_{i}(x_{i},\y)+\sum_{(i,j)\in\edges|i,j\in\tau}E_{ij}(x_{i},x_{j})+\\
 &  & +\sum_{j\in\neighbours(\tau)|(i,j)\in\mathcal{E}}E_{ij}(x_{i},x_{j})+\sum_{(i,j)\in\edges|i,j\notin\tau}E_{ij}(x_{i},x_{j})
\end{eqnarray*}
where 
\begin{itemize}
\item $\sum_{i\in\tau}E_{i}(x_{i},\y)$ is the data energy belonging to
the tree $\tau$,
\item $\sum_{i\notin\tau}E_{i}(x_{i},\y)$ the data energy outside $\tau$, 
\item $\sum_{(i,j)\in\edges|i,j\in\tau}E_{ij}(x_{i},x_{j})$ is the interaction
energy within the tree,
\item $\sum_{j\in\neighbours(\tau)|(i,j)\in\mathcal{E}}E_{ij}(x_{i},x_{j})$
the interaction energy between the tree and its boundary, and
\item $\sum_{(i,j)\in\edges|i,j\notin\tau}E_{ij}(x_{i},x_{j})$ the interaction
energy outside the tree
\end{itemize}
By grouping energies related to the tree and the rest, we have
\begin{eqnarray*}
E(\x_{\tau},\x_{\neg\tau},\y) & = & E_{\tau}\left(\x_{\tau},\xb_{\neighbours(\tau)},\y\right)+\sum_{(i,j)\in\edges|i,j\notin\tau}E_{ij}(x_{i},x_{j})
\end{eqnarray*}
where $E_{\tau}\left(\x_{\tau},\xb_{\neighbours(\tau)},\y\right)$
is given in Eq.~(\ref{eq:conditional-tree-energy}) for all $i\in\tau$.
This leads to:

\begin{eqnarray*}
E(\hat{\x}_{\tau},\x_{\neg\tau},\y) & = & E_{\tau}\left(\hat{\x}_{\tau},\xb_{\neighbours(\tau)},\y\right)+\sum_{(i,j)\in\edges|i,j\notin\tau}E_{ij}(x_{i},x_{j})\\
 & \le & E_{\tau}\left(\x_{\tau},\xb_{\neighbours(\tau)},\y\right)+\sum_{(i,j)\in\edges|i,j\notin\tau}E_{ij}(x_{i},x_{j})\\
 & = & E(\x_{\tau},\x_{\neg\tau},\y)
\end{eqnarray*}
This completes the proof $\blacksquare$

\end{document}